\documentclass[sn-apa]{sn-jnl}%

\jyear{2023}%

\theoremstyle{thmstyleone}%

\theoremstyle{thmstyletwo}%
\newtheorem{remark}{Remark}%

                    \theoremstyle{thmstylethree}%

            \usepackage{lipsum}                     %
            \usepackage{xargs}                      %
            \usepackage[colorinlistoftodos,prependcaption,textsize=tiny]{todonotes}

            \usepackage{blindtext}
            \usepackage{hyperref}
            \usepackage{commath}
            \usepackage[normalem]{ulem} %
            \usepackage[mathscr]{eucal} %
            \usepackage{booktabs} %
            \usepackage{amsmath}
            \usepackage{amsfonts}
            \usepackage{times} %
            \usepackage{ulem} %
            \usepackage[british]{babel} %
            \usepackage[autostyle]{csquotes} %
            \usepackage{enumitem}
            \usepackage{color}
            \usepackage{xcolor}
            
            \usepackage{hyperref}
            
            \usepackage{tikz,tikz-3dplot}%
            
            \tikzset{>=latex}
            \usepackage{pgfplots} %
            \pgfplotsset{compat=1.13}
            \usetikzlibrary{decorations.pathmorphing,decorations.pathreplacing,decorations.shapes}
            \usetikzlibrary{shapes,arrows}
            \usetikzlibrary{positioning,arrows.meta,quotes}
            \usetikzlibrary{shapes,snakes}
            \usetikzlibrary{bayesnet}
            \tikzset{>=latex}

            \ifpdf
            \DeclareGraphicsExtensions{.eps,.pdf,.png,.jpg}
            \else
            \DeclareGraphicsExtensions{.eps}
            \fi
            \usepackage{array} %

            \definecolor{plotColorNearestNeighbor}{rgb}{0.0, 0.5, 0.0}
            \definecolor{plotColorLinearSystem}{rgb}{0.8, 0.6, 0.0}
            \definecolor{plotColorGCN}{rgb}{1.0, 0.0, 0.3}
            \definecolor{plotColorAllenCahn}{rgb}{0.1, 0.6, 0.9}

            \definecolor{darkgreen}{rgb}{0.0, 0.5, 0.1}
            \definecolor{amber}{rgb}{0.9, 0.5, 0.0}

            \newcommand{\SD}[1]{\textcolor{black}{#1}}

            \usepackage{mathptmx}      %

            \usepackage{mymacros}

\begin{document}

            \title[A weighted subspace exponential kernel for support tensor machines]{A weighted subspace exponential kernel for support tensor machines}

            \author*[1]{\fnm{Kirandeep} \sur{Kour}}\email{kour@mpi-magdeburg.mpg.de}
            \equalcont{These authors contributed equally to this work.}

            \author[2]{\fnm{Sergey} \sur{Dolgov}}\email{S.Dolgov@bath.ac.uk}

            \author[1,3]{\fnm{Peter} \sur{Benner}}\email{benner@mpi-magdeburg.mpg.de}

            \author[3]{\fnm{Martin} \sur{Stoll}}\email{martin.stoll@mathematik.tu-chemnitz.de}

            \author[3]{\fnm{Max} \sur{Pfeffer}}\email{max.pfeffer@mathematik.tu-chemnitz.de}
            \equalcont{These authors contributed equally to this work.}

            \affil*[1]{\orgname{Max Planck Institute for Dynamics of Complex Technical Systems}, \orgaddress{\street{Sandtorstra{\ss}e -1}, \city{Magdeburg}, \postcode{39106}, \country{Germany}}}

            \affil[2]{\orgdiv{Department of Mathematical Sciences}, \orgname{University of Bath}, \orgaddress{\city{Bath}, \postcode{BA2 7AY}, \country{United Kingdom}}}

            \affil[3]{\orgdiv{Department of Mathematics}, \orgname{TU Chemnitz}, \orgaddress{\street{ Reichenhainer Str. 41}, \city{Chemnitz}, \postcode{09126}, \country{Germany}}}

\abstract{High-dimensional data in the form of tensors are challenging for kernel classification methods.
To both reduce the computational complexity and extract informative features, kernels based on low-rank tensor decompositions have been proposed.
However, what decisive features of the tensors are exploited by these kernels is often unclear.
In this paper we propose a novel kernel that is based on the Tucker decomposition. For this kernel the Tucker factors are computed based on re-weighting of the Tucker matrices with tuneable powers of singular values from the HOSVD decomposition. This provides a mechanism to balance the contribution of the Tucker core and factors of the data.
We benchmark support tensor machines with this new kernel on several datasets. First we generate synthetic data where two classes differ in either Tucker factors or core, and compare our novel and previously existing kernels. We show robustness of the new kernel with respect to both classification scenarios. We further test the new method on real-world datasets.
The proposed kernel has demonstrated a higher test accuracy than the state-of-the-art tensor train multi-way multi-level kernel, and a significantly lower computational time.

}

\keywords{Kernel methods, Low-rank Tensor Decompositions, Supervised Learning}

\maketitle

\section{Introduction}\label{sec1}
Support Vector Machines (SVM)~\citep{vapnik,vapnik98}, also known as Support Vector Network, maximum margin classifier, are popular machine learning methods, which allow for soft margins and high-dimensional feature embedding by using kernels. However, the standard SVM~\citep{vapnik95} is based on vectors, and may struggle (in terms of both computational complexity and overfitting) for multi-dimensional (tensor) data. Many applications (e.g. in healthcare or signal processing) contain multidimensional data, hence studying kernel methods for tensorial data is an important topic that is addressed in this paper.

The approximation of tensors based on low-rank decompositions has received a lot of attention in scientific computing over recent years~\citep{cichocki2016tensor,Kolda09,Cichocki13,Liu15}.
The tensor-based SVM was introduced as Supervised Tensor Learning (STL) in \cite{Tao07,Zhou13,Guo12, Hao13}.
Using low-rank tensor approximations such as the Canonical Polyadic (CP)~\citep{Hitchcock1927,cp_als}, Tucker~\citep{TuckerLathauwer, tucker1966}, and Tensor Train formats~\citep{oseledets2011tensor, ttcross} within STL alleviates the \emph{curse of dimensionality}, and allows one to reduce both computational complexity, by computing existing kernels faster,
and overfitting, by designing new dedicated kernels using directly the components of the low-rank decomposition \citep{Signoretto11,Signoretto12,QZhao13a}.

In the context of kernel methods, the \textit{Dual Structure-preserving Kernel} (DuSK) for STL, which is particularly tailored to SVM and tensor data, was introduced in~\cite{DuSK}. This kernel is defined using the CP format.
Later, further kernelization in factors, specifically the \textit{Kernelized-CP} (KCP) factorization, have been introduced~\citep{MMK,KSTM17}, and the entire technique has been called the \textit{Multi-way Multi-level Kernel} (MMK) method.
Once an accurate CP approximation is available, DuSK and MMK typically deliver an accurate and efficient classification.
However, the CP approximation of arbitrary data can be numerically unstable and difficult to compute~\citep{desilva2008}. In general, any optimization method (Newton, Steepest Descent or Alternating Least Squares) to obtain the CP decomposition might return only a locally optimal solution, and it is difficult to assess whether this is a local or global optimum.

In contrast, the Tucker approximation problem is well-posed, and a quasi-optimal Tucker approximation can be computed reliably by a sequence of singular value decompositions (SVD) \citep{TuckerLathauwer}.
Therefore, the Tucker format is also used often in data science.
In~\cite{kotsia} the authors have adopted the Tucker decomposition of the weight parameter to retain more structural information, and \cite{Zeng17} extended this by using a Genetic Algorithm (GA) prior to the Support Tucker Machine (STuM) for the contraction of the input feature tensor. In \cite{wolf}, the authors proposed to minimize the rank of the weight parameter with the orthogonality constraints on the columns of the weight parameter instead of the classical maximum-margin criterion, and in~\cite{Pirisiavash} the orthogonality constraints are relaxed to further improve Wolf\textquotesingle s method.

Further understanding of the KCP approach~\cite{KSTM17} is provided by a kernelized Tucker model, inspired by~\cite{Signoretto2013}.

The Tensor Train (TT) decomposition offers a stable approximation similarly to the Tucker format, whereas scaling to higher dimensions like the CP format.	A straightforward generalization of DuSK (MMK) to the TT format was proposed in~\cite{chen2018support}.

However, why exactly the kernels based on low-rank decompositions are good for classification remains unclear.
Moreover, since any tensor decomposition is a nonlinear parametrization of the tensor, its representation may be not unique.
For example, Tucker and TT decompositions are invariant to rotation and scaling of the factors.
These formats can also be converted from one to another, albeit with a change of ranks.
Eliminating redundancy in rotation, scaling, and TT to CP conversion in the TT-MMK method has significantly improved the classification accuracy \cite{KourJMLR}.

               Further attempts to understand the key features of tensorial data and design the kernel accordingly include the Tucker \textit{subspace kernel} \cite{ZhaoKTD2013}.
               	Here, the kernel compares projectors onto the subspaces spanned by Tucker factors.
               	The latter are known to effectively capture the multilinear structure of the data. For example, \cite{Taguchi2021} used HOSVD for unsupervised feature extraction. Multiway analysis enables one to effectively capture the multilinear structure of the data, which is usually available as a priori information about the data. In \cite{FaceReco2007} a subspace learning technique for Face Recognition was introduced.           
The factor match score, a consistent way of comparing the feature vectors of tensor decompositions, has been introduced in~\cite{Acar2011AllatonceOF}.
               
However, the data may contain decisive features not only in the Tucker subspaces, but also in the Tucker core. How to capture both in a computationally efficient way remained largely an open problem. This paper aims to fill this gap by introducing a novel kernel that show high robustness with respect to where the classification information are contained within the tensor.

             \subsection*{Novel contributions}
                The main aim of this paper is to introduce a novel kernel that shows high robustness with respect to where the classification information is contained within the tensor. The main contributions with this novel kernel are summarized as follows:
                \begin{itemize}
                    \item We propose a new form of writing the Tucker (HOSVD) decomposition with weighted factors.
                    \item Based on this form, we propose a new kernel for support tensor machines, which admits fast computation,
                    whereas the weighting takes into account both Tucker factors and core in building the nonlinear decision boundary.
                    \item Using synthetic data with class assignment based on either Tucker factors or core, we confirm that the new kernel provides an accurate classification in all cases, in contrast to existing Tucker-based kernels.
                    \item Finally, we test that the new kernel provides higher classification accuracy than state-of-the-art methods also on real datasets.
                \end{itemize}

\section{Notation and background} \label{preface}
This section sets up and extends notations for tensors and the binary classification problem to multi-dimensional data.
                \subsection{Tensor Algebra}
                In context of numerical multi-linear algebra, a multidimensional array is called a tensor. Tensors are a generalization of matrices (2-modes; rows and columns) with a higher number of \emph{dimensions} / \emph{modes}. We denote all tensors by a calligraphic letter $\tensx$. We assume that all tensors are real-valued. For a general introduction to tensors and their properties we refer to \cite{Kolda09} and the references mentioned therein. We summarize the common notations encountered in this paper in Table~\ref{tab:notation}.

\begin{table}[ht]
\begin{center}
\begin{small}
\begin{tabular}{lllr}
\toprule
Symbol & Description & Definition\\
\midrule
$\overline{i_1,\ldots,i_M}$ & multi-index & $\overline{i_1,\ldots,i_M} = 1 + \sum_{k=1}^{M}(i_k-1) \prod_{m=1}^{k-1} I_m$ \\

$x$    &  scalar value & $x \in \R$\\

\bx  &  vector &  $\bx \in \R^{I}$\\

\bX    &  matrix &  $\bX \in \R^{I \times J}$\\

\tensx    &  tensor & $\tensx \in \dims R I M$\\

$\tenem x$ & $(i_1,i_2,\ldots,i_M)$-entry of a tensor & \\

\nten \tensx m & $m$-mode matricization & $(x_{(m)})_{i_m,\overline{i_1,\ldots,i_{m-1},i_{m+1},\ldots,i_M}}= x_{i_1,\ldots,i_M}$\\

$\tensx \nmod m \bA$  & $m$-mode product& $\nten {(\tensx \nmod m  \bA)} m = \bA \nten \tensx m,$ \ $\bA \in$ \matdim R \\

$\tenz = \tensx \circ \tensy$  & outer product &  $z_{i_1, \ldots, i_{M}, j_1, \ldots, j_N} =  x_{i_1,\ldots,i_M} y_{j_1, \ldots, j_N}.$\\

$\bA \otimes \bB$      & Kronecker product & $
                                        \begin{bmatrix}
                                        a_{i,1} \bB & \cdots & a_{i,J} \bB\\
                                        \end{bmatrix} \in \RIJKL$ \\
                                        
& &                                     $\bA \in \RIJ, \bB \in \RKL$ \\
                                        
$\langle M \rangle$ & integer values from 1 to $M$ & $\{1,2, \cdots, M\}$\\

$\langle \tensx, \tensy \rangle$      & inner product of tensors \tensx \ and \tensy & $ \teninner.$\\

$\| \tensx \|$ & Frobenius norm of the tensor $\tensx$ & $\sqrt{\langle \tensx, \tensx \rangle}$ \\
\bottomrule
\end{tabular}
\end{small}
\end{center}
\caption{Tensor notation used in this paper.}
\label{tab:notation}
\vskip -0.15in
\end{table}

\subsection{Support Tensor Machines for supervised learning}\label{STL}

Although tensor objects can be reshaped into vectors, the structural information encoded in the tensorial data are lost. For example, in an fMRI image, the values of adjacent voxels are typically close to each other~\citep{DuSK}.
It is natural to replace the vector-valued SVM by a tensor-valued SVM  (cf. Supervised Tensor Learning (STL) introduced in \cite{Tao07,Zhou13,Guo12}. An extension of the STL using kernelized tensor factorization with maximum-margin criterion (SVM) is given in \cite{KSTM17}. This preserves the nonlinear structure and enhances the overall performance of the STL model, called Kernelized Support Tensor Machines (KSTM). 
    
\subsubsection{Kernelized Support Tensor Machine}\label{kstm}

                The KSTM is a binary classification model for $N$ tensor input data points $\lbrace \left( \tensx_i, y_i\right) \rbrace _ {i = 1}^N$ 
                where each tensor is of the form $\tensx_i \in \dims R I M$ with labels $y_i \in \{0,1\}$ leading to a nonlinear decision boundary.
                The method follows a maximum margin approach to get the separation hyperplane. Hence, the objective function for a nonlinear boundary can be written as follows~\citep{Cai06supporttensor}:
                \begin{align}\label {eq: tenPrimOpt}
                &\underset{w,b}{\text{min}} \quad \frac{1}{2} \norm{w}^2 + C \sum_{i=1}^{N} \xi_i \\  \nonumber
                \text{subject to} \quad & y_i (\langle \Psi(\tensx_i), w \rangle + b) \geq 1-\xi_{i} ~~ \quad \xi_{i}~\geq 0 ~~~ \forall i.
                \end{align}
                    
                The classification setup given in~\eqref{eq: tenPrimOpt} is known as Support Tensor Machine (STM)~\citep{Tao05}. The dual formulation of the corresponding primal problem is given as follows:
                \begin{align}\label {eq: tenDualOpt}
                &\underset{\alpha_1,\ldots,\alpha_{N}}{\text{max}} \quad  \sum_{i = 1}^{N} \balpha_i - \frac{1}{2} \sum_{i = 1}^{N}\sum_{j = 1}^{N} \balpha_i \balpha_j y_i y_j \langle \Psi(\tensx_i),\Psi(\tensx_j) \rangle \nonumber\\
                &  \text{subject to} \quad 0 \leq \balpha_i \leq C,  \quad \sum_{i = 1}^{N} \balpha_i y_i = 0 ~~ \forall i. 
                \end{align}

                The nonlinear feature embedding for tensor inputs in a tensor space to a feature space is analogous to working with vector inputs in a vector space. We can define an embedding from low-dimensional tensor-product space to the tensor-product Reproducing Kernel Hilbert Space~\citep{KSTM17}. And by using Mercer's Theorem, i.e. having a kernel that is positive semidefinite, we can construct a feature embedding $\Psi$ such that,
                \begin{align*}
                \Psi : \dims R I M \text{ (input space)}  &\rightarrow \mathbb{F}  \text{ (feature space)}\\
\exists~ K \colon \dims R I M \times \dims R I M &\mapsto \R ~~~\text{s.t.}~~~ K\left(\tensx,\tensx'\right) = \langle \Psi(\tensx), \Psi(\tensx') \rangle_{\mathbb{F}}.
                \end{align*}

This is not only computationally tractable but also avoids the explicit computation of the function $\Psi$. The kernel matrix that results from the continued evaluation of the kernel function on the data points is then positive semidefinite. With the help of the kernel, a linear learning algorithm can learn a  \emph{nonlinear boundary}, without explicitly knowing the nonlinear function $\Psi$. 

                Therefore, by using the kernel trick, KSTM is defined as follows:
                \begin{align} \label{eq: kerSTM}
                &\underset{\alpha_1,\ldots,\alpha_{N}}{\text{max}} \quad  \sum_{i = 1}^{N} \balpha_i - \frac{1}{2} \sum_{i = 1}^{N}\sum_{j = 1}^{N} \balpha_i \balpha_j y_i y_j K(\tensx_i,\tensx_j) \nonumber \\
                &  \text{subject to} \quad 0 \leq \balpha_i \leq C, \quad  \sum_{i = 1}^{N} \balpha_i y_i = 0 ~~ \forall i.
                \end{align}
                Once we have the real-valued function (kernel) value for each pair of tensors, we can use the state-of-the-art LIBSVM implementation~\citep{libsvm} , which relies on the \emph{Sequential Minimal Optimization} algorithm to optimize the weights $\alpha_i$. Hence, the preeminent part is the kernel function $K(\tensx_i,\tensx_j)$.

                The STM classifier for predicting labels for unseen test data in tensor form is then given by
                \begin{align}\label{eq:classifier}
                G (\tensx) & = \text{sign} \left( \sum_{i = 1}^{N} \balpha_i y_i K(\tensx_i,\tensx) + b_{0} \right),
                \end{align}
where 
\begin{align}\label{eq:compbval}
                b_0 &=\frac{1}{N_0} \sum_{i: \balpha_i\in(0,C)} \left(y_i-\sum_{j=1}^{N} \balpha_j K(\tensx_j, \tensx_i) \right), ~~~ \text{with} ~~~ N_0= \sum_{i: \alpha_i \in (0,C)} 1 .
                \end{align}
The only task needed for the KSTM is thus to choose a well-suited kernel function. This way, we can work with the input data in a high-dimensional space, while all computations are performed in the original low-dimensional space. We will discuss possible choices for tensor kernels in Sec.~\ref{sec:tkernels}, where we will also introduce a novel tensor kernel. These kernels are based on low-rank formats for tensors, which we introduce now.
                
\subsection{Low-rank Tensor Decompositions}
                Given the complexity of storing the full tensor $\tensx$, it is often desirable to have a different potentially more economic representation.  As such, tensor decomposition methods have seen much progress over the last two decades, and they are applied to solve problems of varying computational complexity. The main goal is the linear (or at most polynomial) scaling of the computational complexity in the dimension (order) of a tensor. The key ingredient is the separation of variables via approximate low-rank factorizations. 

\begin{figure}[ht]
                            \begin{center}
                                \centerline{\hspace{-2cm}%
                \scalebox{1}{\definecolor{mpiblue}{HTML}{33a5c3}
\colorlet{MPIblue}{mpiblue}
\definecolor{mpibluefont}{HTML}{17a1c1}
\colorlet{MPIbluefont}{mpibluefont}
\definecolor{mpigreen}{HTML}{007675}
\colorlet{MPIgreen}{mpigreen}
\definecolor{mpired}{HTML}{78004B}
\colorlet{MPIred}{mpired}
\definecolor{mpisand}{HTML}{ece9d4}
\colorlet{MPIsand}{mpisand}
\newcommand{\Depth}{1.8}
\newcommand{\Height}{1.5}
\newcommand{\Width}{1.2}
\newcommand{\xx}{1}
\newcommand{\yy}{1}
\newcommand{\zz}{1}
\begin{tikzpicture}
 \coordinate (O) at (0,0,0);
\coordinate (A) at (0,\Width,0);
\coordinate (B) at (0,\Width,\Height);
\coordinate (C) at (0,0,\Height);
\coordinate (D) at (\Depth,0,0);
\coordinate (E) at (\Depth,\Width,0);
\coordinate (F) at (\Depth,\Width,\Height);
\coordinate (G) at (\Depth,0,\Height);
\shadedraw[inner color=white,outer color=mpiblue,draw=black] (O) -- (C) -- (G) -- (D) -- cycle;%
\shadedraw[inner color=white,outer color=mpiblue,draw=black,opacity = 0.6] (O) -- (A) -- (E) -- (D) -- cycle;%
\shadedraw[inner color=white,outer color=mpiblue,draw=black, opacity = 0.6]  (O) -- (A) -- (B) -- (C) -- cycle;%
\shadedraw[inner color=white,outer color=mpiblue,draw=black, opacity = 0.5](D) -- (E) -- (F) -- (G) -- cycle;%
\shadedraw[inner color=white,outer color=mpiblue,draw=black,opacity=0.5]  (C) -- (B) -- (F) -- (G) -- cycle;%
\shadedraw[inner color=white,outer color=mpiblue,draw=black,opacity=0.5] (A) -- (B) -- (F) -- (E) -- cycle;%

\draw [line width = 0.5pt,decorate,decoration={brace,amplitude=5pt,mirror,raise=1pt},xshift=10pt,yshift=1pt]
 (B) -- (C) node[above left= 0.4cm and 0.1cm]
{\footnotesize {\scalebox{0.7}{$I_1$}}}; 
\draw [line width = 0.5pt,decorate,decoration={brace,amplitude=5pt,mirror,raise=1pt},xshift=10pt,yshift=1pt]
 (C) -- (G) node[below left= 0.15cm and 0.7cm]
{\footnotesize {\scalebox{0.7}{$I_2$}}}; 
\draw [line width = 0.5pt,decorate,decoration={brace,amplitude=5pt,mirror,raise=1pt},xshift=10pt,yshift=1pt]
 (G) -- (D) node[below right= 0.35cm and -0.2cm]
{\footnotesize {\scalebox{0.7}{$I_3$}}}; 

 \coordinate (O) at (0+\xx,0+0.7\yy,0+\zz);  %
 \coordinate (F) at (1.5+\xx,0 +0.7\yy, 0+\zz);

\draw (O) node {$\tensx$}; %
\draw (F) node {$\cong$}; %

    \draw [very thick] (2.5,-0.7) rectangle (2.7,0.3);
    \filldraw [fill=mpired!40!white,draw=mpigreen!40!black] (2.5,-0.7) rectangle (2.7,0.3);
    \draw (3,-0.6) node {\scriptsize{$\ba^{(1)}_1$}};
   \draw [very thick] (2.8, 0.4) rectangle (4,0.6);
   \filldraw [fill=mpigreen!40!white,draw=mpigreen!40!black] (2.8,0.4) rectangle (4,0.6);
   \draw (3.8, 0.15) node {\scriptsize{$\ba^{(2)}_1$}};
    \draw[fill=olive!40!white,draw=mpigreen!40!black, thick] (2.5,0.7) --(2.7,0.7)--(3.2,1.3)--(3,1.3) -- cycle;
   \draw (3.5,1.3) node {\scriptsize{$\ba^{(3)}_1$}};
   
    \draw (4.4,0.5) node {{\color{black}\large{$+$}}};
    \draw (6.5,0.5) node {{\color{black}\large{$+$}}};
    \draw (7,0.5) node {{\color{black}\large{$\cdots$}}};
    \draw (7.5,0.5) node {{\color{black}\large{$+$}}};
    \draw [very thick] (4.7,-0.7) rectangle (4.9,0.3);
    \filldraw [fill=mpired!40!white,draw=mpigreen!40!black] (4.7,-0.7) rectangle (4.9,0.3);
    \draw (5.2,-0.6) node {\scriptsize{$\ba^{(1)}_2$}};
   \draw [very thick] (5, 0.4) rectangle (6.2,0.6);
   \filldraw [fill=mpigreen!40!white,draw=mpigreen!40!black] (5,0.4) rectangle (6.2,0.6);
   \draw (6, 0.15) node {\scriptsize{$\ba^{(2)}_2$}};
    \draw[fill=olive!40!white,draw=mpigreen!40!black, thick] (4.7,0.7) --(4.9,0.7)--(5.4,1.3)--(5.2,1.3) -- cycle; 
   \draw (5.7,1.3) node {\scriptsize{$\ba^{(3)}_2$}};

    \draw [very thick] (7.7,-0.7) rectangle (7.9,0.3);
    \filldraw [fill=mpired!40!white,draw=mpigreen!40!black] (7.7,-0.7) rectangle (7.9,0.3);
    \draw (8.2,-0.6) node {\scriptsize{$\ba^{(1)}_R$}};
   \draw [very thick] (8, 0.4) rectangle (9.2,0.6);
   \filldraw [fill=mpigreen!40!white,draw=mpigreen!40!black] (8,0.4) rectangle (9.2,0.6);
   \draw (9, 0.15) node {\scriptsize{$\ba^{(2)}_R$}};
    \draw[fill=olive!40!white,draw=mpigreen!40!black, thick] (7.7,0.7) --(7.9,0.7)--(8.4,1.3)--(8.2,1.3) -- cycle; 
   \draw (8.7,1.3) node {\scriptsize{$\ba^{(3)}_R$}};

\end{tikzpicture}}  
            }
                                \caption{CP decomposition of a 3-way tensor of rank $R$.}
                                \label{figcp}
                            \end{center}
                            \vskip -0.2in
                            \end{figure}
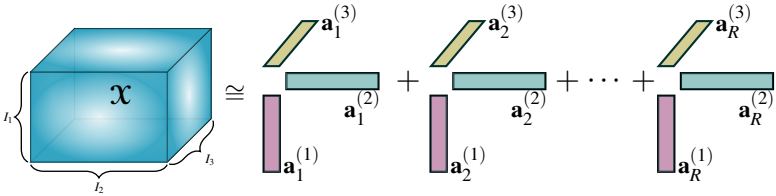

                            \paragraph{Canonical Polyadic decomposition}\label{subsec:cp}
                            The Canonical Polyadic (CP) decomposition of an $M^{th}-$order tensor $\tensx \in \dims R I M$ is a factorization into a sum of rank-one components~\citep{Hitchcock1927}, which is given element-wise as
                            \begin{align}\label{eq:cpkrush}
                            \tenem x & \cong \sum_{r=1}^{R} a^{(1)}_{i_1,r} a^{(2)}_{i_2,r} \cdots a^{(M)}_{i_M,r},  \nonumber\\
                            \mbox{or shortly,} ~~\qquad \quad \quad \tensx & \cong \llbracket \cpd \rrbracket,
                            \end{align}
                            where  $\mathbf{A}^{(m)} = \left[ a^{(m)}_{i_m,r} \right] \in \R^{I_m \times R}$, $m=1,\ldots,M$, are called \textit{factor matrices} of the CP decomposition, see~Fig.~\ref{figcp}, and $R$ is called the CP-rank. The notation $\llbracket \cpd \rrbracket$ is also called the Kruskal representation of the tensor.
                            Despite the simplicity of the CP format, the problem of the best CP approximation is often ill-posed~\citep{desilva2008}. 
                            A practical CP approximation can be computed via the Alternating Least Squares (ALS) method~\citep{cp_als}, but convergence may be slow.
                            It may also be difficult to choose the rank $R$.

                            \paragraph{Tucker Decomposition}\label{subsec:tucker}

                            The Tucker decomposition consists of a decomposition of the tensor into matrices and a core tensor, where the core tensor has smaller dimension compared to the original tensor.  For a given tensor $ \tensx \in \dims R I M$ the Tucker decomposition is as then given by,
                            \begin{align}\label{eq:tucker}
                            \tensx & \cong \sum_{r_1 = 1}^{R_1} \sum_{r_2 = 1}^{R_2} \ldots \sum_{r_M= 1}^{R_M} g_{r_1 r_2 \ldots r_M} \left( \bu_{r_1}^{(1)} \circ \bu_{r_2}^{(2)} \circ \ldots \circ \bu_{r_M}^{(M)}\right) \nonumber \\
                            & = \llbracket \teng ;  \bU^{(1)}, \bU^{(2)}, \ldots, \bU^{(M)} \rrbracket . \\
                            & \nonumber
                            \end{align}
                            \vskip -0.2in
                            \begin{figure}[ht]
                            \begin{center}
                                \centerline{%
                \scalebox{0.8}{ 
\definecolor{mpiblue}{HTML}{33a5c3}
\colorlet{MPIblue}{mpiblue}
\definecolor{mpibluefont}{HTML}{17a1c1}
\colorlet{MPIbluefont}{mpibluefont}
\definecolor{mpigreen}{HTML}{007675}
\colorlet{MPIgreen}{mpigreen}
\definecolor{mpired}{HTML}{78004B}
\colorlet{MPIred}{mpired}
\definecolor{mpisand}{HTML}{ece9d4}
\colorlet{MPIsand}{mpisand}
\newcommand*\circledb[1]{\tikz[baseline=(char.base)]{
            \node[shape=circle,draw,inner sep=2pt,fill = mpiblue] (char) {#1};}}
 \newcommand*\circled[1]{\tikz[baseline=(char.base)]{
            \node[shape=circle,draw,inner sep=2pt,fill = mpired!50] (char) {#1};}}    
\scalebox{0.05}{
\begin{tikzpicture}
	\coordinate (A1) at (0em,0cm,0cm); %
	\coordinate (A2) at (1.2cm,0cm,0cm);
	\coordinate (A3) at (0em,0.8cm,0cm); %
	\coordinate (A4) at (0em,0cm,0.9cm);
	\coordinate (A5) at (1.2cm,0.8cm,0cm);
        \coordinate (A6) at (1.2cm,0cm,0.9cm);
        \coordinate (A7) at (0em,0.8cm,0.9cm);
        \coordinate (A8) at (1.2cm,0.8cm,0.9cm);
        
	\draw[thick] (A3) -- (A5);
	\draw[thick] (A3) -- (A7);
	 \draw [line width = 15pt,decorate,decoration={brace,amplitude=70pt},xshift=300pt,yshift=5pt]
         (A2) -- (A6) node[above right = -1cm and 8cm]
        {\footnotesize {\scalebox{18}{$R_3$}}};
	 \draw [line width = 15pt,decorate,decoration={brace,amplitude=70pt,mirror,raise=4pt},xshift=300pt,yshift=5pt]
         (A4) -- (A6) node[below left = 4cm and 15cm]{\footnotesize {\scalebox{18}{$R_2$}}}; %
	\draw[thick] (A8) -- (A5);
	\draw[thick] (A8) -- (A7);
	\draw[thick] (A8) -- (A6);
        \draw [line width = 15pt,decorate,decoration={brace,amplitude=70pt},xshift=300pt,yshift=5pt]
        (A4) -- (A7) node[below left = 10cm and 3cm]
       {\footnotesize {\scalebox{18}{$R_1$}}}; %
	\draw[thick,dashed] (A1) -- (A2) ;
 	\draw[thick,dashed] (A1) -- (A4); %
 	\draw[thick,dashed] (A1) -- (A3);

	\shadedraw[inner color=white,outer color=mpigreen,draw=black] (A1) -- (A2) -- (A5) -- (A3) -- cycle; %
        \shadedraw[inner color=white,outer color=mpigreen,draw=black,opacity = 0.6] (A1) -- (A4) -- (A7) -- (A3) -- cycle; %
        \shadedraw[inner color=white,outer color=mpigreen,draw=black, opacity = 0.6] (A2) -- (A5) -- (A8) -- (A6) -- cycle; %
        \shadedraw[inner color=white,outer color=mpigreen,draw=black, opacity = 0.5] (A4) -- (A6) -- (A8) -- (A7) -- cycle; %
	\shadedraw[inner color=white,outer color=mpigreen,draw=black,opacity=0.2] (A1) -- (A2) -- (A6) -- (A4) -- cycle; %
	\shadedraw[inner color=white,outer color=mpigreen,draw=black,opacity=0.2] (A3) -- (A5) -- (A8) -- (A7) -- cycle; %
	\node (G) at (barycentric cs:A4=1,A6=1,A8=1,A7=1) {};
        \node (G1) [right =  2 cm] at (G) {\scalebox{25}{$\boldsymbol{\mathscr{G}}$}};
         
	\coordinate (B5) at (-1cm,0cm,2cm);
        \coordinate (B6) at (-0.2cm,0cm,2cm);
        \coordinate (B7) at (-1cm,1.2cm,2cm);
        \coordinate (B8) at (-0.2cm,1.2cm,2cm);
       
	 \draw [line width = 15pt,decorate,decoration={brace,amplitude=70pt,mirror,raise=4pt},xshift=300pt,yshift=5pt]
         (B5) -- (B6) node[below left = 4cm and 8cm]{\footnotesize {\scalebox{18}{$R_1$}}}; %
	 \draw [line width = 15pt,decorate,decoration={brace,amplitude=70pt},xshift=300pt,yshift=5pt]
        (B5) -- (B7) node[below left = 15cm and 3cm]
        {\footnotesize {\scalebox{18}{$I_1$}}};
	\draw[thick] (B8) -- (B7); %
	\draw[thick] (B8) -- (B6); %
		
	\shadedraw[inner color=white,outer color=mpired!70,draw=black] (B6) -- (B8) -- (B7) -- (B5) -- cycle; %
	\node (C1) at (barycentric cs:B6=1,B8=1,B7=1,B5=1) {};
       \node (C1') [] at (C1) {\scalebox{15}{$\boldsymbol{U}^{(1)}$}};
         
	\coordinate (C5) at (1.8cm,0cm,0cm);
        \coordinate (C6) at (3cm,0cm,0cm);
        \coordinate (C7) at (1.8cm,0.7cm,0cm);
        \coordinate (C8) at (3cm,0.7cm,0cm);
       
	 \draw [line width = 15pt,decorate,decoration={brace,amplitude=70pt,mirror,raise=4pt},xshift=300pt,yshift=5pt]
         (C5) -- (C6) node[below left = 4cm and 14cm]{\footnotesize {\scalebox{18}{$I_2$}}}; %
	 \draw [line width = 15pt,decorate,decoration={brace,amplitude=70pt},xshift=300pt,yshift=5pt]
        (C8) -- (C6) node[above right = 7cm and 3cm]
        {\footnotesize {\scalebox{18}{$R_2$}}};
	\draw[thick] (C8) -- (C7); %
	\draw[thick] (C5) -- (C7); %
		
	\shadedraw[inner color=white,outer color=orange!60,draw=black] (C6) -- (C8) -- (C7) -- (C5) -- cycle; %
	\node (D1) at (barycentric cs:C6=1,C8=1,C7=1,C5=1) {};
       \node (D1') [] at (D1) {\scalebox{15}{$\boldsymbol{U}^{(2)}$}};
         
        \coordinate (x1) at (-4.1cm,-0.3cm,0cm); %
	\coordinate (x2) at (-2.7cm,-0.3cm,0cm);
	\coordinate (x3) at (-4.1cm,1.1cm,0cm); %
	\coordinate (x4) at (-4.1cm,-0.3cm,1.4cm);
	\coordinate (x5) at (-2.7cm,1.1cm,0cm);
        \coordinate (x6) at (-2.7cm,-0.3cm,1.4cm);
        \coordinate (x7) at (-4.1cm,1.1cm,1.4cm);
        \coordinate (x8) at (-2.7cm,1.1cm,1.4cm);
        
	\draw[thick] (x3) -- (x5);
	\draw[thick] (x3) -- (x7);
	\draw[thick] (x2) -- (x6); 
	\draw[thick] (x4) -- (x6);
	\draw[thick] (x8) -- (x5);
	\draw[thick] (x8) -- (x7);
	\draw[thick] (x8) -- (x6);
	\draw[thick] (x4) -- (x7); 
	\draw[thick,dashed] (x1) -- (x2) ;
 	\draw[thick,dashed] (x1) -- (x4); %
 	\draw[thick,dashed] (x1) -- (x3);

	\shadedraw[inner color=white,outer color=mpiblue,draw=black] (x1) -- (x2) -- (x5) -- (x3) -- cycle; %
	
 	\shadedraw[inner color=white,outer color=mpiblue,draw=black,opacity = 0.6] (x1) -- (x4) -- (x7) -- (x3) -- cycle; %
	
	\shadedraw[inner color=white,outer color=mpiblue,draw=black, opacity = 0.6] (x2) -- (x5) -- (x8) -- (x6) -- cycle; %
	
	\shadedraw[inner color=white,outer color=mpiblue,draw=black, opacity = 0.5] (x4) -- (x6) -- (x8) -- (x7) -- cycle; %
	
	\shadedraw[inner color=white,outer color=mpiblue,draw=black,opacity=0.5] (x1) -- (x2) -- (x6) -- (x4) -- cycle; %
	
	\shadedraw[inner color=white,outer color=mpiblue,draw=black,opacity=0.5] (x3) -- (x5) -- (x8) -- (x7) -- cycle; %
	\node (X1) at (barycentric cs:x3=1,x5=1,x8=1,x7=1) {};
	\node (X1') [below = 17cm] at (X1) {\scalebox{25}{$\boldsymbol{\mathscr{X}}$}};

        \draw [line width = 15pt,decorate,decoration={brace,amplitude=70pt,mirror,raise=4pt},xshift=300pt,yshift=5pt]
 (x4) -- (x6) node[below left = 4cm and 18cm]
{\footnotesize {\scalebox{18}{$I_2$}}}; 
        \draw [line width = 15pt,decorate,decoration={brace,amplitude=70pt},xshift=300pt,yshift=5pt]
 (x4) -- (x7) node[below left = 18cm and 3cm]
{\footnotesize {\scalebox{18}{$I_1$}}};
        \draw [line width = 15pt,decorate,decoration={brace,amplitude=70pt},xshift=300pt,yshift=5pt]
 (x2) -- (x6) node[above right = 3cm and 11cm]
{\footnotesize {\scalebox{18}{$I_3$}}};

        \node [inner sep=0pt,below right = 12cm and 32cm] at (X1) {\scalebox{30}{$\cong$}};

	\coordinate (D5) at (1cm,2.4cm,1.5cm);
        \coordinate (D6) at (2cm,2.4cm,1.5cm);
        \coordinate (D7) at (0.7cm,2cm,2cm);
        \coordinate (D8) at (1.7cm,2cm,2cm);
       
	 \draw [line width = 15pt,decorate,decoration={brace,amplitude=70pt,mirror,raise=4pt},xshift=300pt,yshift=5pt]
         (D7) -- (D8) node[below left = 4cm and 10cm]{\footnotesize {\scalebox{18}{$R_3$}}}; %
	 \draw[thick]  (D5) -- (D7);
       
	\draw[thick] (D5) -- (D6); %
	\draw [line width = 15pt,decorate,decoration={brace,amplitude=70pt},xshift=300pt,yshift=5pt] (D6) -- (D8) node[above right = 2cm and 10cm] {\footnotesize {\scalebox{18}{$I_3$}}}; %
		
	\shadedraw[inner color=white,outer color=brown!70,draw=black] (D6) -- (D8) -- (D7) -- (D5) -- cycle; %
	\node (E1) at (barycentric cs:D6=1,D8=1,D7=1,D5=1) {};
       \node (E1') [] at (E1) {\scalebox{15}{$\boldsymbol{U}^{(3)}$}};
        
\end{tikzpicture}
}}  
            }
                                \caption{Tucker decomposition of a  3-way tensor.}
                                \label{figtucker}
                            \end{center}
                            \vskip -0.2in
                            \end{figure}
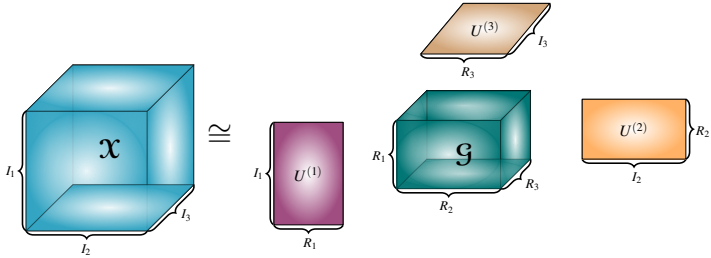

Here, $\teng$ is a tensor of size $\R^{R_{1} \times R_{2} \times \cdots \times R_{M}}$ and $R_{m}$ is the Tucker rank in each mode matricization of the tensor $\tensx$. A crucial advantage of the Tucker format (and all tree tensor networks) is the ability to perform algebraic operations directly on the component tensors, avoiding full tensors. Moreover, we can compute a quasi-optimal Tucker approximation of any given tensor using the SVD. This builds on the fact that the Tucker decomposition constitutes a successive matrix factorization, where each Tucker rank is the matrix rank of the appropriate unfolding of the tensor, and hence the Tucker approximation problem is well-posed \citep{TuckerLathauwer}. One unique advantage of the Tucker format is the interpretability of the leaf-components $\bU^{(m)}$: Since they result directly from an SVD of the $m$-th matricization, their columns constitute an orthonormal basis of the subspace of $\mathbb R^{I_m}$ that the data lies in.

\begin{figure}[ht]
                            \begin{center}
                                \centerline{ %
                \scalebox{0.04}{\definecolor{mpiblue}{HTML}{33a5c3}
\colorlet{MPIblue}{mpiblue}
\definecolor{mpibluefont}{HTML}{17a1c1}
\colorlet{MPIbluefont}{mpibluefont}
\definecolor{mpigreen}{HTML}{007675}
\colorlet{MPIgreen}{mpigreen}
\definecolor{mpired}{HTML}{78004B}
\colorlet{MPIred}{mpired}
\definecolor{mpisand}{HTML}{ece9d4}
\colorlet{MPIsand}{mpisand}
\newcommand*\circledb[1]{\tikz[baseline=(char.base)]{
            \node[shape=circle,draw,inner sep=2pt,fill = mpiblue] (char) {#1};}}
 \newcommand*\circled[1]{\tikz[baseline=(char.base)]{
            \node[shape=circle,draw,inner sep=2pt,fill = mpired!50] (char) {#1};}}    

\begin{tikzpicture}
	\coordinate (A1) at (0em,0cm,0cm); %
	\coordinate (A2) at (1.4cm,0cm,0cm);
	\coordinate (A3) at (0em,0.8cm,0cm); %
	\coordinate (A4) at (0em,0cm,0.9cm);
	\coordinate (A5) at (1.4cm,0.8cm,0cm);
        \coordinate (A6) at (1.4cm,0cm,0.9cm);
        \coordinate (A7) at (0em,0.8cm,0.9cm);
        \coordinate (A8) at (1.4cm,0.8cm,0.9cm);
        
	\draw[thick] (A3) -- (A5);
	\draw[thick] (A3) -- (A7);
	\draw[thick] (A2) -- (A6);
	\draw[thick] (A4) -- (A6); %
	\draw[thick] (A8) -- (A5);
	\draw[thick] (A8) -- (A7);
	\draw[thick] (A8) -- (A6);
	\draw[thick] (A4) -- (A7); %
	\draw[thick,dashed] (A1) -- (A2) ;
 	\draw[thick,dashed] (A1) -- (A4); %
 	\draw[thick,dashed] (A1) -- (A3);

	\fill[mpired!70] (A1) -- (A2) -- (A5) -- (A3) -- cycle; %
	
 	\fill[mpired!50,opacity = 0.6] (A1) -- (A4) -- (A7) -- (A3) -- cycle; %
	
	\fill[mpired!50, opacity = 0.6] (A2) -- (A5) -- (A8) -- (A6) -- cycle; %
	
	\fill[mpired!30, opacity = 0.5] (A4) -- (A6) -- (A8) -- (A7) -- cycle; %
	
	\fill[mpigreen!50,opacity=0.2] (A1) -- (A2) -- (A6) -- (A4) -- cycle; %
	
	\fill[mpigreen!90,opacity=0.2] (A3) -- (A5) -- (A8) -- (A7) -- cycle; %
	\node (G2) at (barycentric cs:A3=1,A5=1,A8=1,A7=1) {};
 	\node (G2') [below = 10cm] at (G2) {\scalebox{20}{$\boldsymbol{\tendtt C 2}$}};

            \coordinate (B1) at (-2.4cm,0.2cm,0cm); %
            \coordinate (B2) at (-1cm,0.2cm,0cm);
            \coordinate (B3) at (-2.4cm,0.5cm,0cm); %
            \coordinate (B4) at (-2.4cm,0.2cm,0.8cm);
            \coordinate (B5) at (-1cm,0.5cm,0cm);
            \coordinate (B6) at (-1cm,0.2cm,0.8cm);
            \coordinate (B7) at (-2.4cm,0.5cm,0.8cm);
            \coordinate (B8) at (-1cm,0.5cm,0.8cm);
            
        \draw[thick] (B3) -- (B5);
        \draw[thick] (B3) -- (B7);
        \draw[thick] (B2) -- (B6); %
        \draw[thick] (B4) -- (B6);%
        \draw[thick] (B8) -- (B5);
        \draw[thick] (B8) -- (B7);
        \draw[thick] (B8) -- (B6);
        \draw[thick] (B4) -- (B7); %
        \draw[thick,dashed] (B1) -- (B2) ;
        \draw[thick,dashed] (B1) -- (B4); %
        \draw[thick,dashed] (B1) -- (B3);

	\fill[mpired!70] (B1) -- (B2) -- (B5) -- (B3) -- cycle; %
	
 	\fill[mpired!50,opacity = 0.6] (B1) -- (B4) -- (B7) -- (B3) -- cycle; %
	
	\fill[mpired!50, opacity = 0.6] (B2) -- (B5) -- (B8) -- (B6) -- cycle; %
	
	\fill[mpired!30, opacity = 0.5] (B4) -- (B6) -- (B8) -- (B7) -- cycle; %
	
	\fill[mpigreen!50,opacity=0.2] (B1) -- (B2) -- (B6) -- (B4) -- cycle; %
	
	\fill[mpigreen!90,opacity=0.2] (B3) -- (B5) -- (B8) -- (B7) -- cycle; %
	\node (G1) at (barycentric cs:B3=1,B5=1,B8=1,B7=1) {};
 	\node (G1') [below =2cm] at (G1) {\scalebox{20}{$\boldsymbol{\tendtt C 1}$}};

    \coordinate (C1) at (2.6cm,0.1cm,0cm); %
	\coordinate (C2) at (2.9cm,0.1cm,0cm);
	\coordinate (C3) at (2.6cm,1cm,0cm); %
	\coordinate (C4) at (2.6cm,0.1cm,1.4cm);
	\coordinate (C5) at (2.9cm,1cm,0cm);
    \coordinate (C6) at (2.9cm,0.1cm,1.4cm);
    \coordinate (C7) at (2.6cm,1cm,1.4cm);
    \coordinate (C8) at (2.9cm,1cm,1.4cm);
        
	\draw[thick] (C3) -- (C5);
	\draw[thick] (C3) -- (C7);
	\draw[thick] (C2) -- (C6);%
	\draw[thick] (C4) -- (C6); %
	\draw[thick] (C8) -- (C5);
	\draw[thick] (C8) -- (C7);
	\draw[thick] (C8) -- (C6);
	\draw[thick] (C4) -- (C7); %
	\draw[thick,dashed] (C1) -- (C2) ;
 	\draw[thick,dashed] (C1) -- (C4) ;
 	\draw[thick,dashed] (C1) -- (C3);

	\fill[mpired!70] (C1) -- (C2) -- (C5) -- (C3) -- cycle; %
	
 	\fill[mpired!50,opacity = 0.6] (C1) -- (C4) -- (C7) -- (C3) -- cycle; %
	
	\fill[mpired!50, opacity = 0.6] (C2) -- (C5) -- (C8) -- (C6) -- cycle; %
	
	\fill[mpired!30, opacity = 0.5] (C4) -- (C6) -- (C8) -- (C7) -- cycle; %
	
	\fill[mpigreen!50,opacity=0.2] (C1) -- (C2) -- (C6) -- (C4) -- cycle; %
	
	\fill[mpigreen!90,opacity=0.2] (C3) -- (C5) -- (C8) -- (C7) -- cycle; %
	\node (G3) at (barycentric cs:C3=1,C5=1,C8=1,C7=1) {};
	\node (G3') [below right= 10cm and -1.5cm] at (G3) {\scalebox{20}{$\tendtt C 3$}};
        \draw [line width = 15pt,decorate,decoration={brace,amplitude=70pt},xshift=300cm,yshift=5cm]
        (B5) -- (B8) node (k1) [black,midway,xshift=1.2cm, yshift = -1.25cm]  {}; %
        \draw [line width = 15pt,decorate,decoration={brace,amplitude=70pt,mirror,raise=4pt},xshift=300pt,yshift=5pt]
        (A7) -- (A4) node (k2) [black,midway,xshift=-5.5cm] %
        {\footnotesize \scalebox{15}{$\boldsymbol{R_1}$}}; %
        \draw [line width = 15pt,decorate,decoration={brace,amplitude=70pt},xshift=300cm,yshift=5cm]
        (A5) -- (A8) node (k3) [black,midway,xshift=1.2cm, yshift = -0.8cm]  {};
        \draw [line width = 15pt,decorate,decoration={brace,amplitude=70pt,mirror,raise=4pt},xshift=300cm,yshift=5cm]
        (C7) -- (C4) node (k4) [black,midway,xshift=-5.5cm] 
        {\footnotesize \scalebox{15}{$\boldsymbol{R_2}$}}; %
        \draw[-latex,line width=3mm,dotted] (k1) -- (k2) ;
        \draw[-latex,line width=3mm,dotted] (k3) -- (k4) ;
        \draw [line width = 15pt,decorate,decoration={brace,amplitude=70pt,mirror,raise=4pt},xshift=300pt,yshift=5pt]
        (B4) -- (B6) node[ below left = 3.5cm and 17cm]
        {\footnotesize {\scalebox{18}{$I_1$}}};   %
        \draw [line width = 15pt,decorate,decoration={brace,amplitude=70pt,mirror,raise=4pt},xshift=300pt,yshift=5pt]
        (A4) -- (A6) node[below left =4cm and  17cm]
        {\footnotesize {\scalebox{18}{$I_2$}}}; %
        \draw [line width = 15pt,decorate,decoration={brace,amplitude=70pt},xshift=300pt,yshift=5pt]
        (C2) -- (C6) node[above right = 2cm and 10cm ,opacity = 4.9]
        {\footnotesize {\scalebox{18}{$I_3$}}};       %
            
        \coordinate (x1) at (-5.1cm,-0.2cm,0cm); %
        \coordinate (x2) at (-3.7cm,-0.2cm,0cm);
        \coordinate (x3) at (-5.1cm,1.2cm,0cm); %
        \coordinate (x4) at (-5.1cm,-0.2cm,1.4cm);
        \coordinate (x5) at (-3.7cm,1.2cm,0cm);
        \coordinate (x6) at (-3.7cm,-0.2cm,1.4cm);
        \coordinate (x7) at (-5.1cm,1.2cm,1.4cm);
        \coordinate (x8) at (-3.7cm,1.2cm,1.4cm);
            
        \draw[thick] (x3) -- (x5);
        \draw[thick] (x3) -- (x7);
        \draw[thick] (x2) -- (x6); 
        \draw[thick] (x4) -- (x6);
        \draw[thick] (x8) -- (x5);
        \draw[thick] (x8) -- (x7);
        \draw[thick] (x8) -- (x6);
        \draw[thick] (x4) -- (x7); 
        \draw[thick,dashed] (x1) -- (x2) ;
        \draw[thick,dashed] (x1) -- (x4); %
        \draw[thick,dashed] (x1) -- (x3);

        \shadedraw[inner color=white,outer color=mpiblue,draw=black] (x1) -- (x2) -- (x5) -- (x3) -- cycle; %
        
        \shadedraw[inner color=white,outer color=mpiblue,draw=black,opacity = 0.6] (x1) -- (x4) -- (x7) -- (x3) -- cycle; %
        
        \shadedraw[inner color=white,outer color=mpiblue,draw=black, opacity = 0.6] (x2) -- (x5) -- (x8) -- (x6) -- cycle; %
        
        \shadedraw[inner color=white,outer color=mpiblue,draw=black, opacity = 0.5] (x4) -- (x6) -- (x8) -- (x7) -- cycle; %
        
        \shadedraw[inner color=white,outer color=mpiblue,draw=black,opacity=0.5] (x1) -- (x2) -- (x6) -- (x4) -- cycle; %
        
        \shadedraw[inner color=white,outer color=mpiblue,draw=black,opacity=0.5] (x3) -- (x5) -- (x8) -- (x7) -- cycle; %
        \node (X1) at (barycentric cs:x3=1,x5=1,x8=1,x7=1) {};
        \node (X1') [below = 17cm] at (X1) {\scalebox{25}{$\boldsymbol{\mathscr{X}}$}};

        \draw [line width = 15pt,decorate,decoration={brace,amplitude=70pt,mirror,raise=4pt},xshift=300pt,yshift=5pt]
        (x4) -- (x6) node[below left = 4cm and 18cm]
        {\footnotesize {\scalebox{18}{$I_2$}}}; 
        \draw [line width = 15pt,decorate,decoration={brace,amplitude=70pt},xshift=300pt,yshift=5pt]
        (x4) -- (x7) node[below left = 18cm and 3cm]
        {\footnotesize {\scalebox{18}{$I_1$}}};
        \draw [line width = 15pt,decorate,decoration={brace,amplitude=70pt},xshift=300pt,yshift=5pt]
        (x2) -- (x6) node[above right = 2cm and 10cm]
        {\footnotesize {\scalebox{18}{$I_3$}}};

        \node [inner sep=0pt,below right = 14cm and 35cm] at (X1) {\scalebox{30}{$\cong$}};
        
        \node (P1) at (barycentric cs:B2=1,B5=1,B6=1,B8=1) {};
        \node (P2) at (barycentric cs:A1=1,A4=1,A7=1,A3=1) {};
        \node (P3) at (barycentric cs:A2=1,A5=1,A8=1,A6=1) {};
        \node (P4) at (barycentric cs:C1=1,C4=1,C7=1,C3=1) {};

\end{tikzpicture}}  
            }
                                \caption{TT decomposition of a 3-way tensor.}
                                \label{figtt}
                            \end{center}   
                            \vskip -0.2in
                            \end{figure}
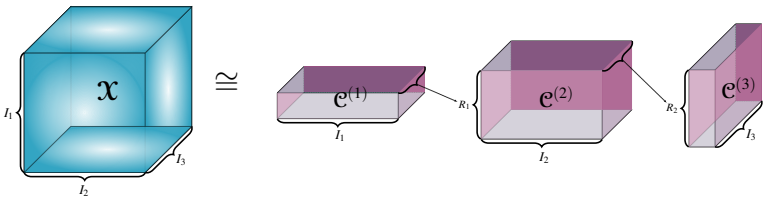

                            \paragraph{Tensor Train decomposition}\label{subsec:tt}
                            The Tensor Train (TT)~\citep{oseledets2011tensor} decomposition of an $M^{th}-$order tensor  $\tensx \in \dims R I M$ is defined element-wise as
                            \begin{align}\label{eq:tt}
                            \tenem x & \cong \sum_{r_0,\ldots,r_{M}}c_{r_0,i_1,r_1}^{(1)} c_{r_1,i_2,r_2}^{(2)} \cdots c_{r_{M-1},i_M,r_M}^{(M)}, \nonumber\\
                            \tensx & \cong\llangle \tendtt C 1, \tendtt C 2, \ldots, \tendtt C M \rrangle,
                            \end{align}
                            where \tendtt C m $\in$ \ttcdim m, $m=1,\ldots,M,$ are 3rd-order tensors called \emph{TT-cores} (see~Fig.~\ref{figtt}), and $R_0,\ldots,R_M$ with $R_0 = R_M = 1$ are called \emph{TT-ranks}.
                            Since the TT decomposition is also a tree-based tensor format, all above-mentioned advantages of the Tucker format also translate to the TT format. Furthermore, complexity of the TT format is quadratic in the ranks, whereas the Tucker format scales exponentially with the Tucker ranks. This reduction of complexity however comes at the cost of interpretability: There is no straightforward way to interpret the meaning of the TT components.

                            \subsubsection{Converting Tucker and TT into CP}\label{sec:convcp}

                            It is easy to see from~\eqref{eq:cpkrush},~\eqref{eq:tucker}, and~\eqref{eq:tt} that one can convert a tensor in Tucker or TT format into the CP format without too much effort: Summing over all Tucker ranks $R_1,\ldots,R_M$ or all TT-ranks $r_0,\ldots,r_M$ will yield a sum of rank-one tensors as required in the CP format. We emphasize here that this does not result in a {\em minimal} CP decomposition but only in a CP {\em representation} of the tensor. This can however still be useful, as long as the obtained CP representation retains the interpretable qualities of the Tucker or TT decomposition.

                            The main difficulty with obtaining a meaningful conversion is that none of the decomposition formats is really unique: CP allows for a rescaling of columns of the factor matrices, and for Tucker and TT, one can insert identity matrices $I = Q Q^{-1}$ between the modes without changing the tensor. 

                            In a previous work~\citep{KourJMLR}, the problem of meaningfully converting TT into CP was overcome by enforcing uniqueness in the TT-SVD, then converting into CP, and then equilibrating the column norms of the factor matrices in order to avoid ambiguity in the CP representation.

                            This technique can be used analogously for the conversion of Tucker into CP: In the Higher Order SVD (HOSVD), we enforce uniqueness by fixing the sign of each singular vector. This is done by finding the element of maximum absolute value and making it positive. The Tucker tensor is then converted to CP by simply summing over all ranks $R_1,\ldots,R_M$ and a norm equilibration is performed in order to distribute the scalar $g_{r_1r_2\ldots r_M}$ in~\eqref{eq:tucker} across all factors.

\section{Kernels for tensor data in SVM}\label{sec:tkernels}
In this section, we discuss possible choices for tensor kernels. First, we briefly recapitulate existing tensor kernels before we introduce our new kernel. This, together with the numerical study of tensor kernel in Sec.~\ref{sec:synth}, is the main result of this article. At the end of this section, we compare the complexity of computing the different kernels.

\subsection{Existing kernels}

\paragraph{The Gaussian kernel}
The natural idea of defining a kernel for tensorial data would be to extend the classical Gaussian kernel directly from vector to tensor format. That is, the computation can be given directly as follows, 
    \begin{align}
    K(\tensx, \tensy) &= \exp \Big(\frac{- \norm{\tensx - \tensy}_F^2}{2 g^2} \Big),
    \end{align}
with $g$ being the length scale of the kernel. We compute the distance between the two input tensors using the Frobenius norm. This norm can be computed efficiently in each of the tensor formats introduced above  (Table~\ref{tab:complexity} shows the leading terms of the complexity estimates). However, by treating the tensor as a simple vector, we lose valuable information about the different tensor modes. It has been observed (e.g. in~\cite{DuSK,KourJMLR}) that this straightforward idea yields suboptimal classification results and that it can be improved by introducing more sophisticated tensor kernels.

\paragraph{Dual Structure-preserving Kernel}
The Dual Structure-preserving Kernel (\textbf{DuSK}) was introduced first in~\cite{DuSK} for a rank-one tensor factorization and was later extended for the Kernelized CP decomposition in~\cite{MMK}. For given tensors $\tensx \in \dims R I M$ and $\tensy \in \dims R I M$ and their corresponding CP decomposition given by $\llbracket \bA^{(1)}, \bA^{(2)}, \ldots, \bA^{(M)} \rrbracket$ and $\llbracket \bB^{(1)}, \bB^{(2)}, \ldots, \bB^{(M)} \rrbracket$, the formulation of the kernel approximation by DuSK is given as follows:
    \begin{align}\label{eq:kerAPRX}
    \langle \Psi(\tensx), \Psi(\tensy) \rangle & =  K(\tensx,\tensy)\nonumber\\
    & = K \left( \sum_{i=1}^{R} \ba_i^{(1)}\otimes \ba_i^{(2)}\otimes \cdots \otimes \ba_i^{(M)}, \sum_{j=1}^{R} \bb_j^{(1)}\otimes \bb_j^{(2)}\otimes \cdots \otimes \bb_j^{(M)}  \right)\nonumber\\
    & = \sum_{i,j = 1}^{R}\nk(\ba_{i}^{(1)},\bb_{j}^{(1)}) \nk(\ba_{i}^{(2)},\bb_{j}^{(2)}) \cdots \nk(\ba_{i}^{(M)},\bb_{j}^{(M)}),
    \end{align}
where,
\begin{align}
	\nk(\ba,\bb) = \exp \left(\frac{-\norm{\ba-\bb}^2}{2 g^2} \right).
\end{align}
In short, we evaluate the kernel function $\nk(\cdot,\cdot)$ on the individual factors of the CP decomposition. The motivation of DuSK is simple: Since the CP decomposition is often unique (up to norm equilibration), comparing the feature vectors in each mode directly will most likely improve classification. However, this kernel is inherently designed for CP tensors, which is why we have to convert other tensor formats into the CP format first (see Sec.~\ref{sec:convcp} and \cite{KourJMLR}).

\paragraph{The subspace kernel}
Instead of comparing the feature vectors in the CP decomposition, one can use a similar approach for the Tucker format. Here, the feature vectors are stored in the leaf-components $\bU^{(m)}$ and they span the column spaces of the $m$-th matricizations. Thus, it makes sense to compare the {\em projections} onto these subspaces. This kernel was introduced in \cite{ZhaoKTD2013}.

Let $\tensx \in \dims R I M$ denote an $M^{th}$-order tensor, when the SVD is applied on the mode-$m$ unfolding as  $\tensx_{(m)} = \bU_{\tensx}^{(m)} \Sigma_{\tensx}^{(m)} \bV_{\tensx}^{(m)^{\mathsf T}}$ and similarly for $\tensy \in \dims R I M$,  $\tensy_{(m)} = \bU_{\tensy}^{(m)} \Sigma_{\tensy}^{(m)} \bV_{\tensy}^{(m)^{\mathsf T}}$, then the Chordal distance-based kernel is defined as,
    \begin{align}
    K(\tensx,\tensy) &= \prod_{m = 1}^{M} \exp \Big(-{\frac{1}{2 g^2} \norm{ \bU_{\tensx}^{(m)} \bU_{\tensx}^{(m)^{\mathsf T}} - \bU_{\tensy}^{(m)} \bU_{\tensy}^{(m)^{\mathsf T}}}}_F^2 \Big)
    \end{align}
This kernel provides us with rotation and reflection invariance for elements on the Grassmann manifold. Furthermore, the kernel does not see the core tensor $\teng$ and is invariant under rescaling of the feature vectors (i.e., the columns of the leaf matrices).

\subsection{The weighted subspace exponential kernel}\label{subsec:wsek}

The subspace kernel performs well when the information about the classification is stored in the subspaces $\bU_{\tensx}^{(m)}$ and it performs poorly when the information is mostly contained in the core tensor $\teng$. This is confirmed in our synthetic numerical experiments below. The DuSK uses a similar strategy as the subspace kernel: Here, we compare all the feature vectors in the CP decomposition. DuSK therefore also performs better if most of the information is in the subspaces, i.e., in the feature vectors.

The main {contribution} of this article is an improved tensor kernel for Tucker tensors that also includes information of the core tensor $\teng$,  combining the strengths of the subspace kernel and DuSK, and that outperforms both of them in common scenarios, and can be computed more efficiently than DuSK.

We first observe that in the computation of the SVD of a matricized tensor $\tensx_{(m)}$, we can shift any power of the singular values into either the left or the right singular matrices:
\begin{equation*}
	\tensx_{(m)} = \bU^{(m)} \boldsymbol\Sigma^{(m)} (\bV^{(m)})^T = \bU^{(m)} \left( \boldsymbol\Sigma^{(m)} \right)^p \left( \boldsymbol\Sigma^{(m)} \right)^{1-p} (\bV^{(m)})^T
\end{equation*}
\SD{for any $p\in \mathbb{R}$ (assuming no zero singular values, or defining $0^0=1$).}
Using this, we can distribute singular values over the Tucker factors in the HOSVD (see Algorithm~\ref{alg:SqrtmHOSVD}) and we use the resulting {\em rescaled} features $\bar \bU^{(m)} = \bU^{(m)} \left( \boldsymbol\Sigma^{(m)} \right)^p$ for the computation of the kernel. We note that by default, we choose $p = \frac{1}{M}$ as this corresponds to \SD{distributing the norm} $\| \tensx \| = \| \boldsymbol \Sigma^{(m)} \|_F$ \SD{equally over the $M$ feature matrices, which provides accurate classification in practice.}

Since the subspace kernel is invariant under rescaling of the features, we compute the Euclidean distances instead and sum over the exponential kernels of all combinations, noting that if the distances are large, these terms will be negligible. The result is similar to DuSK, but it uses the feature vectors from the Tucker decomposition and the order of the sum over the ranks and the product over the tensor order is reversed:

        \begin{align}\label{eq:rev-wFMS}
        K(\tensx,\tensy) &=   \prod_{m = 1}^{M} \sum_{i,j=1}^{R_m} \nk((\bu_{\tensx}^{(m)})_i, (\bu_{\tensy}^{(m)})_j) = \SD{\prod_{m = 1}^{M}  \sum_{i,j = 1}^{R_m}\exp \left(- \frac{1}{2 g^2}\norm{\bar \bu_{\tensx,i}^{(m)} - \bar \bu_{\tensy,j}^{(m)}}_F^2\right),}
        \end{align}
\SD{where, as in the subspace kernel, we distinguish the SVD of the two tensors, writing $\tensx_{(m)} = \bU_{\tensx}^{(m)} \Sigma_{\tensx}^{(m)} \bV_{\tensx}^{(m)^T}$ and $\tensy_{(m)} = \bU_{\tensy}^{(m)} \Sigma_{\tensy}^{(m)} \bV_{\tensy}^{(m)^T}$.}
We call this kernel the {\em weighted subspace exponential kernel} or WSEK for short. We also considered leaving the order of the sum and the product the same as in DuSK, but this would need further considerations if the Tucker ranks across the modes are not all the same. Furthermore, our experiments have not shown any significant difference in classification accuracy \SD{with respect to the order of $\prod_{m = 1}^{M}$ and $\sum_{i,j=1}^{R_m}$.}

    \begin{algorithm}[t!]
    \caption{Weighted HOSVD}
     \label{alg:SqrtmHOSVD}
    \begin{algorithmic}
        \Require Given tensor  $\tensx \in \dims R I M$, Tucker ranks $R_1,\ldots,R_M$, weighting power $p$ (default $p=1/M$).
        \Ensure Tucker factors $\bar \bU^{(1)},\ldots,\bar \bU^{(M)}$ and core $\teng$.
        \For{$m = 1$ to $M$}
        \State Step 1: \emph{Computing uniqueness-enforced HOSVD}
        \State Compute SVD $\left[ \bU^{(m)}, \boldsymbol\Sigma^{(m)}, (\bV^{(m)})^{\mathsf T} \right] = \text{svd}(\tensx_{(m)})$,
        \State where
        $\boldsymbol\Sigma^{(m)} = \text{diag}(\sigma_1^{(m)}, \sigma_2^{(m)}, \ldots, \sigma_{I_m}^{(m)})$ %
        \For {$r_m = 1$ to $R_m$}
        \State $i_{r_m}^* = \arg\max_{i=1,\ldots,I_m} |u_{i,r_m}^{(m)}|$  \label{abs}
                \State $\mathbf{\hat u}_{r_m}^{(m)}:= \mathbf{u}_{r_m}^{(m)} / \mathrm{sign}(u_{i^*_{r_m},r_m}^{(m)})$
        \EndFor
        \State $\hat \bU^{(m)} = [\mathbf{\hat u}_1^{(m)}, \mathbf{\hat u}_2^{(m)}, \ldots, \mathbf{\hat u}_{R_m}^{(m)}]$ \label{ttcore}
        
        \State Step 2: \emph{Computing norm weighted factors}
        \State $\bar \bU^{(m)} = \hat \bU^{(m)} \boldsymbol\Sigma_p^{(m)}$
        \State where $\boldsymbol\Sigma_p^{(m)} =  \text{diag}((\sigma_1^{(m)})^{p}, (\sigma_2^{(m)})^{p}, \ldots, (\sigma_{R_m}^{(m)})^{p}) $
        \State $\teng \leftarrow \texttt{ttm}(\tensx, \hat \bU^{(m)} (\boldsymbol\Sigma_p^{(m)})^{-1}, m)$ ~~~~ \citep{st_hosvd}
        \EndFor 
    \end{algorithmic}
    \end{algorithm}

\subsection{Computational complexity of the different kernels}\label{subsec:complexity}

For a large number $N$ of data points, for large tensor order $M$ or large mode dimensions $I_m$, computing the different tensor kernels can be very time consuming. If the data input is already given in CP format, computing the DuSK is not too expensive. But most often, the data is given as a full tensor. In these cases, we prefer to compute first a TT or Tucker decomposition of the tensor and then convert it into CP, in order to circumvent the aforementioned numerical issues with the computation of the CP decomposition. Here, we will however only note the complexity of the kernel computation with respect to the given ranks. We denote the maximal dimensions or ranks by $I = \max_{m = 1,\ldots,M} I_m$, $R_{Tucker} = \max_{m = 1,\ldots,M} R_m$, and $R_{TT} = \max_{m = 1,\ldots,M-1} r_m$. The ranks are also to be understood as the maximal respective rank of all data inputs.

                \begin{table}[b]
                    \caption{Theoretical complexity of computing a single entry of the different kernel matrices from data given in different formats. Note that some kernel-format combinations are not defined.}
                    \label{tab:complexity}
                    \begin{center}
                        \begin{tabular}{|c|c|c|c|c|}
                            \hline 
                            & Full & CP & Tucker & TT\\
                            \hline
                            Gaussian & $\mathcal{O}(I^M)$ & $\mathcal{O}(MIR^2)$ & $\mathcal{O}(MR_{Tucker}^{(M+1)} + MIR_{Tucker}^2)$ & $\mathcal{O}(MIR_{TT}^3)$\\
                            \hline
                            DuSK & -- & $\mathcal{O}(MIR^2)$ & $\mathcal{O} (MIR_{Tucker}^{2M})$ & $\mathcal{O} (MIR_{TT}^{2(M-1)})$ \\
                            \hline
                            Subspace & -- & $\mathcal{O}(MIR(I+R))$ & $\mathcal{O}(MI R_{Tucker} (I+R_{Tucker}))$ & -- \\
                            \hline
                            WSEK & -- & $\mathcal{O}(MIR^2)$ & $\mathcal{O}(MIR_{Tucker}^2)$ & -- \\
                            \hline 
                        \end{tabular}
                    \end{center}
                    \vskip 0.15in
                \end{table}

Table~\ref{tab:complexity} summarizes the computational complexity for a single entry of the kernel matrix. We note that all kernels except the Gaussian kernel can only be computed if the tensor is in a low-rank format. Also, we interpret a tensor in CP format to be a Tucker tensor with diagonal core tensor $\teng$ and thus the subspace kernel and WSEK are computed using the factor matrices. Furthermore, in the cases of Tucker and TT tensors, we report on the complexity using naive matrix multiplication ($\mathcal O(n^3)$).

We observe that for large tensor order $M$, computation of the Gauss kernel is prohibitive if it is not done in a low-rank format. If the CP rank $R$ is small, all kernels can be computed efficiently. However, if the CP decomposition has to be obtained by conversion from Tucker or TT, these ranks can be large and DuSK suffers from the curse of dimensionality. WSEK is even more efficient than the subspace kernel. We will report on the CPU times for our numerical experiments in Sec.~\ref{numel}.

\section{A numerical study on synthetic data}\label{sec:synth}

As mentioned above, knowledge about the tensor structure can and should be exploited when choosing the tensor kernel $K$. In this section, we explore why the DuSK performs well in many cases by comparing it to the Gaussian kernel and the Subspace kernel in a synthetic experimental setting. Furthermore, we show that our proposed WSEK retains the advantages of DuSK, while outperforming it in cases where DuSK is less suitable.

                \subsection{Interpreting CP and Tucker}\label{sec:intcptuck}

                The CP decomposition of a tensor can be seen as a special case of the Tucker decomposition with a diagonal core tensor. More precisely, if the matrix ranks of the factor matrices $\mathbf A^{(1)},\ldots,\mathbf A^{(M)}$ are smaller than $R$, we can find a Tucker decomposition of the tensor with Tucker ranks $R_m < R$ for $m = 1,\ldots,M$. The core is then no longer diagonal but it will have many zero entries if $R < R_1 + \cdots + R_M$.

                In any case, for $m = 1,\ldots,M$, the factor matrix $\mathbf A^{(m)}$ in the CP format, or the leaf $\bU^{(m)}$ in the Tucker format, spans the subspace of $\mathbb R^{I_m}$ that the data of the $m$-th mode lies in (the columns of the $m$-mode matricization of the full tensor also span this subspace). The information that is stored in the Tucker tensor (and by inclusion also in the CP tensor) is therefore twofold: What are the subspaces that our data lies in? This information is stored in the orthogonalized leafs of the Tucker tensor. And what combination of feature vectors is present (and to what degree) in the data? This information is stored in the core tensor $\teng$.

                This observation can be exploited when designing a tensor kernel for KSTM. The subspace kernel introduced in Section~\ref{sec:tkernels} can only see the subspaces, i.e.~the leafs of the Tucker tensor. DuSK includes the core data via the norm equilibration. The Gaussian kernel uses the whole tensor but it is less sensible in spotting the subspaces, because the tensor is simply vectorized and this information is hidden. Our proposed WSEK includes information of the core tensor because the feature vectors are weighted with the $p$-th power of the corresponding singular values.

                \subsection{A Synthetic Experiment}\label{sec:synthexp}

                We substantiate our considerations by creating two artificial experimental scenarios: In one (the {\em leaf}-scenario), all the information necessary for classification is stored in the leafs (i.e.~the subspaces) and in the other (the {\em core}-scenario), all the information is in the core of the Tucker tensor. We then test the performance of the aforementioned kernels on this data for different noise levels and tensor ranks.

                The detailed experimental setup is as follows: Let $M = 3$ and $I_1 = I_2 = I_3 = 100$. For different Tucker ranks $R_1 = R_2 = R_3 = r_{approx} = 1,\ldots,10$, we simulate the approximation of a tensor with Tucker ranks $R_1' = R_2' = R_3' = r_{exact} = 3$ plus some noise. That is, the core $\teng$ of the simulated tensor will have size $r_{approx} \times r_{approx} \times r_{approx}$ and the leafs will have sizes $I_m \times r_{approx}$ for $m = 1,2,3$. The core consists of random noise that is normally distributed with mean 0 and variance $\vartheta^2$ (the noise level to be chosen later). To the small upper-left-and-foremost cube of size $\min(r_{approx},3) \times \min(r_{approx},3) \times \min(r_{approx},3)$, we add the information tensor. In the {\em core}-scenario, this is the same tensor for all samples in the same class, generated by drawing the entries from a standard normal distribution. In the {\em leaf}-scenario, the information tensor is different for all samples (also drawn from the standard normal distribution).

                The leafs of size $I_m \times r_{approx}$ also consist of random noise (normally distributed with mean 0 and variance $\vartheta^2$) and to the first $\min(r_{approx},3)$ columns we add vectors $\cos(\pi*\nu*{\mathbf v})$, where $\nu \in \mathbb R^{100}$ is a uniform discretization of $[-1,1]$ and the frequency $\nu$ is chosen uniformly at random (with mean 0 and variance 1). In the {\em leaf}-scenario, these frequencies are the same for all samples in the same class, and in the {\em core}-scenario, these frequencies are different for all samples. After the construction, the leafs are orthogonalized (using the QR-decomposition and discarding the $R$-matrix), so that the resulting Tucker tensor is already in the form of a HOSVD.

                The reasoning is that these two scenarios yield Tucker tensors of rank $r_{approx}$ that are approximations of noisy tensors of rank $r_{exact}$, and the cluster information is stored exclusively in either the core or the leafs. We generate 100 samples in two classes with 50 samples each for each noise level $\vartheta^2 = 0.01,0.02,0.05,0.1,0.2,0.5,1$ and Tucker ranks $r_{approx} = 1,3,5,10$. We then perform the SVM 20 times with 5-fold cross validation in order to determine the hyperparameters $C \in \{2^{-8 :1:8} \}$ (the soft-margin parameter) and $g \in \{2^{-4 :1:12} \}$ (the variance parameter in the Gaussian kernel).

                \begin{figure}[t]
                	\includegraphics[scale = 0.4]{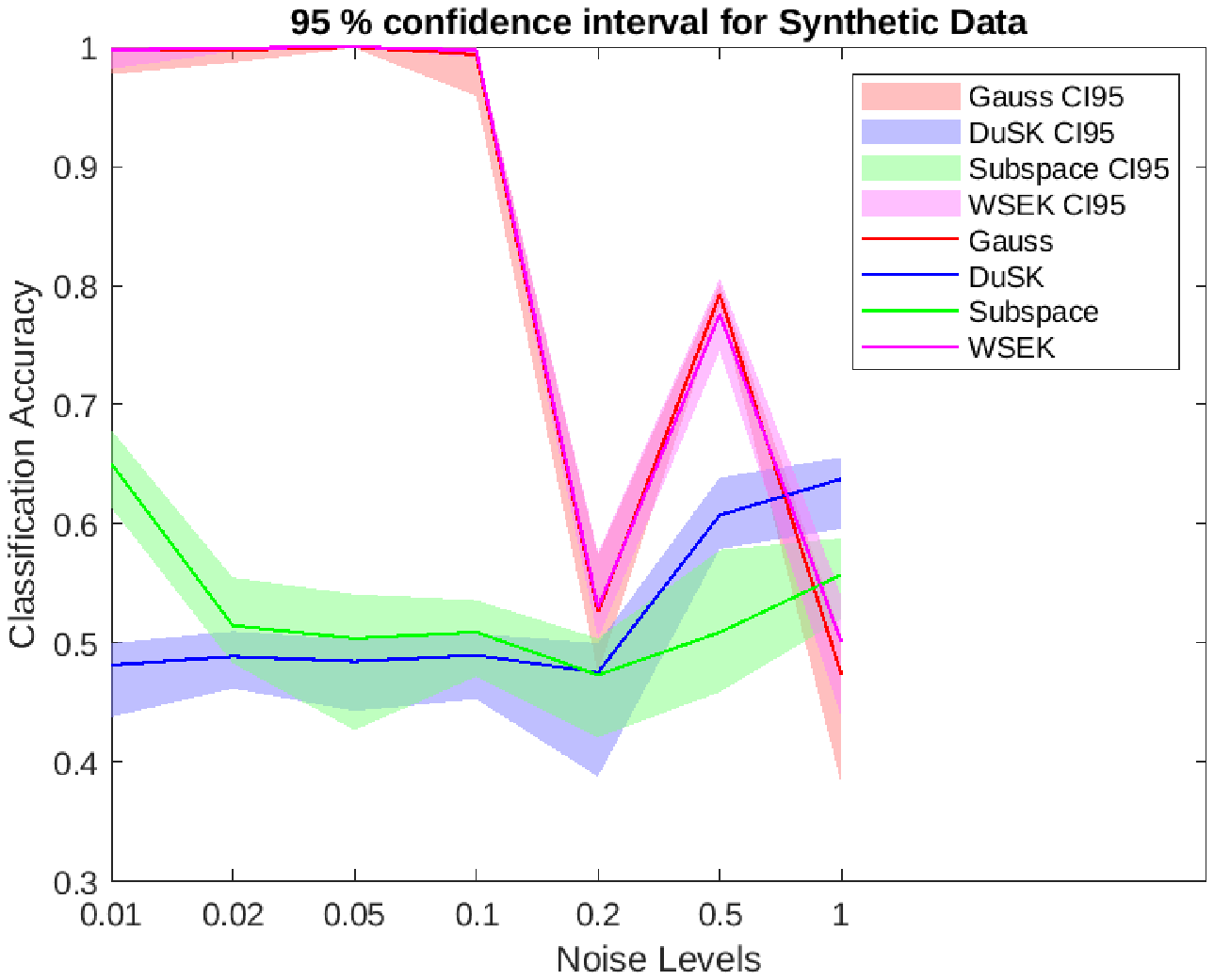}
                	\includegraphics[scale = 0.4]{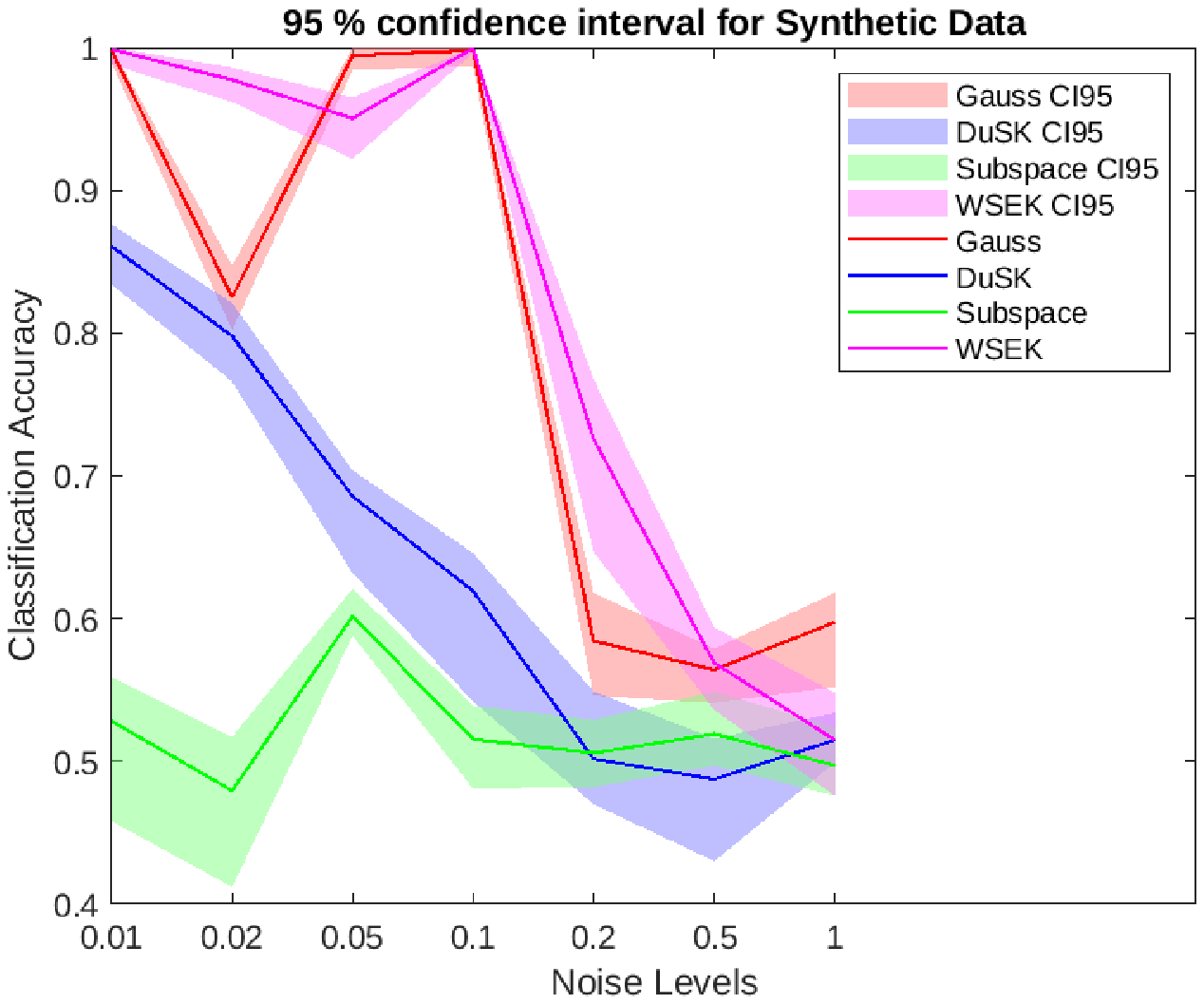}\\
                	\includegraphics[scale = 0.4]{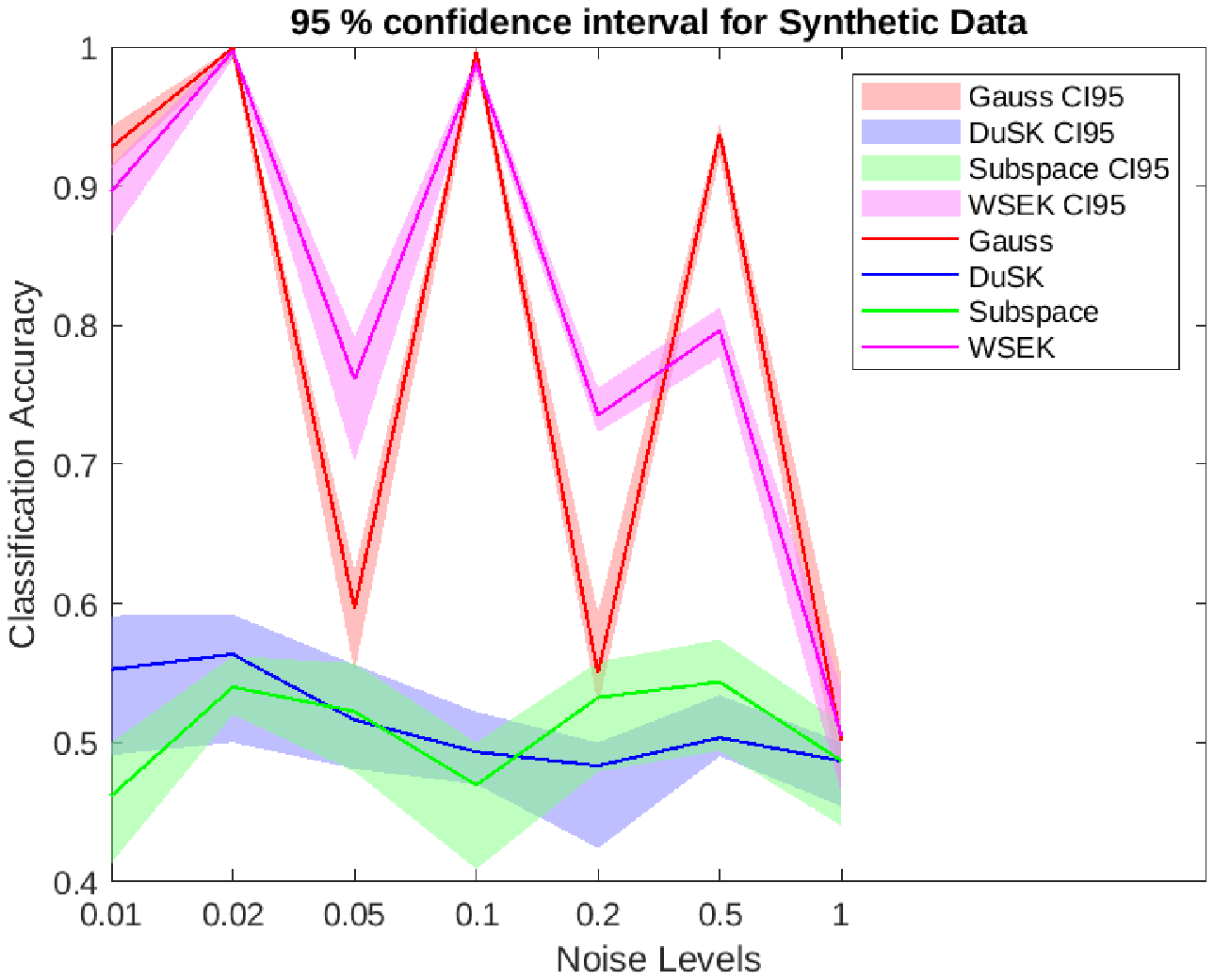}
                	\includegraphics[scale = 0.4]{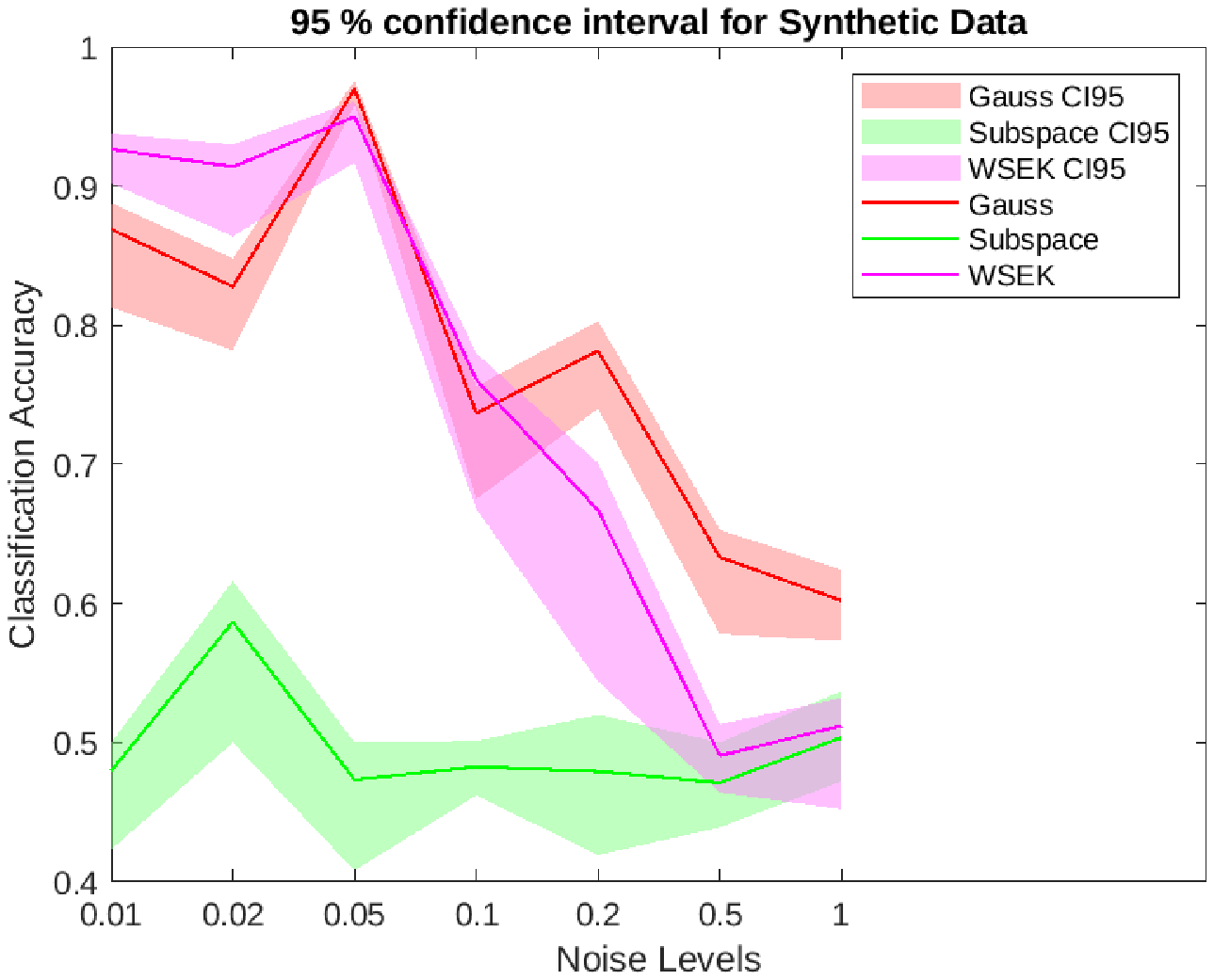}
                	\caption{Classification Accuracy with 95\% confidence interval for different noise levels in the {\em core}-scenario of the generated synthetic data with different rank trucation $r_{approx} = 1,3,5,10$ (DuSK not included for $r_{approx} = 10$)}
                    \label{fig:synthCore}
                \end{figure}

                \subsection{Results and Interpretation}

\begin{figure}[t]
                	\includegraphics[scale = 0.4]{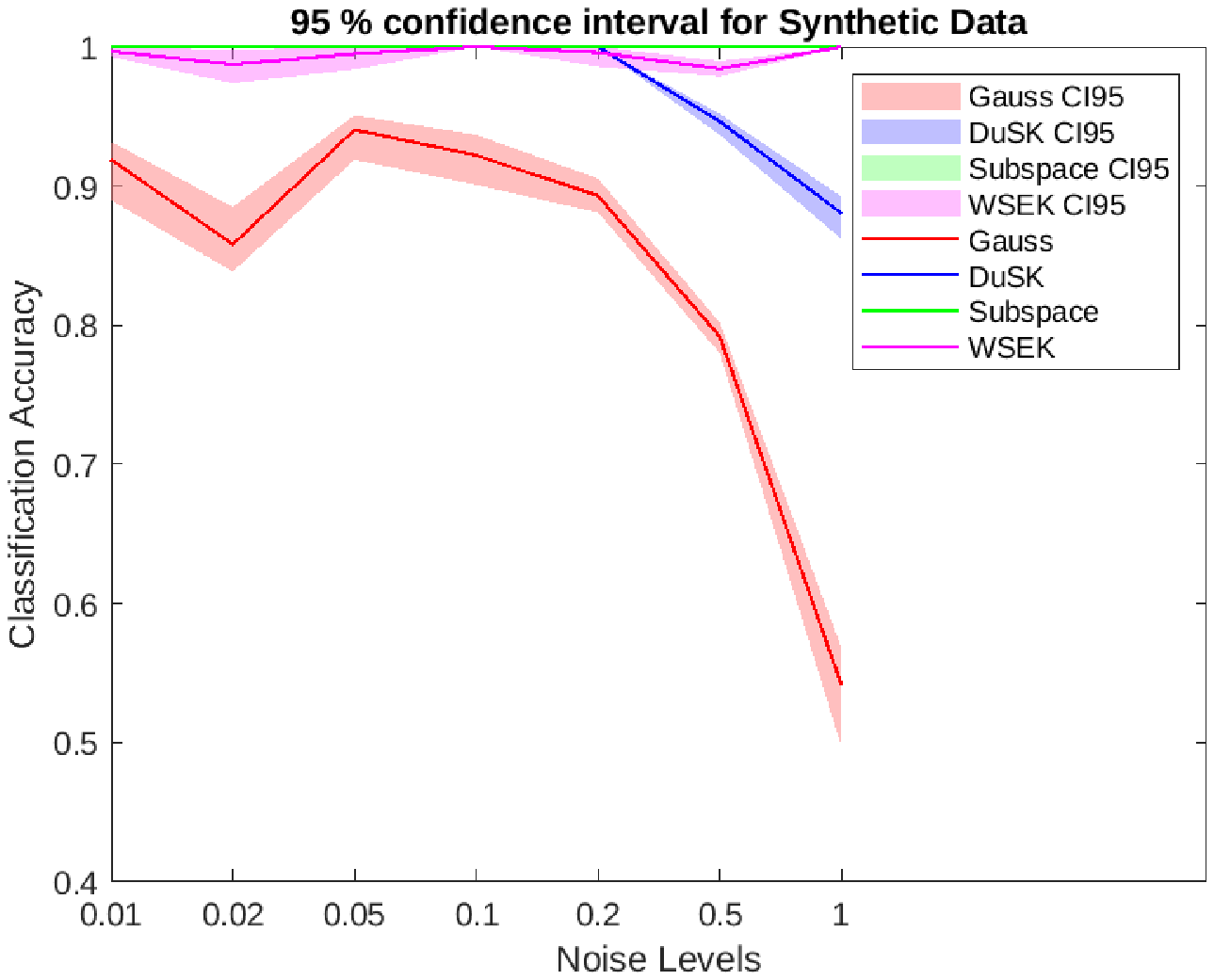}
                	\includegraphics[scale = 0.4]{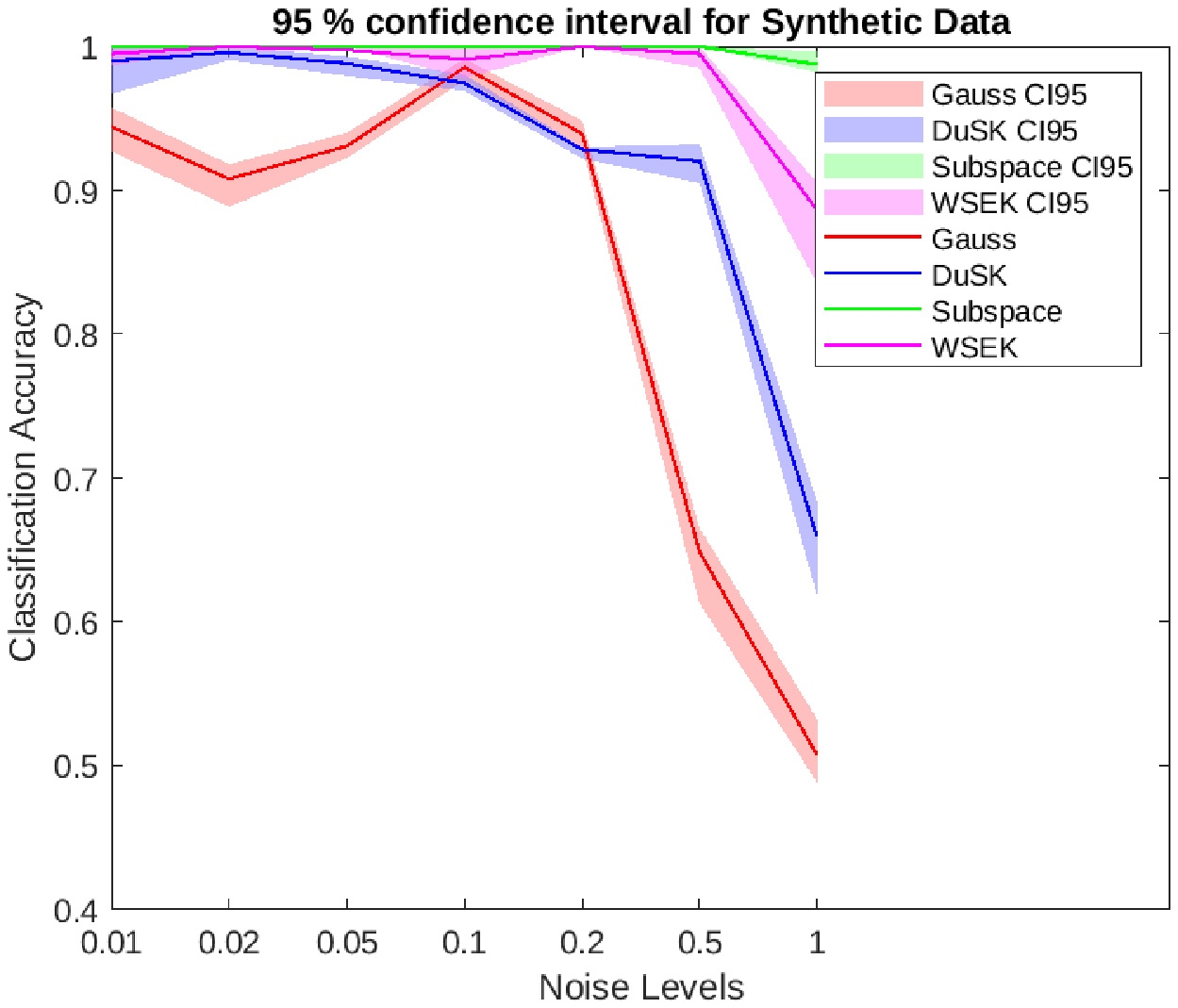}\\
                	\includegraphics[scale = 0.4]{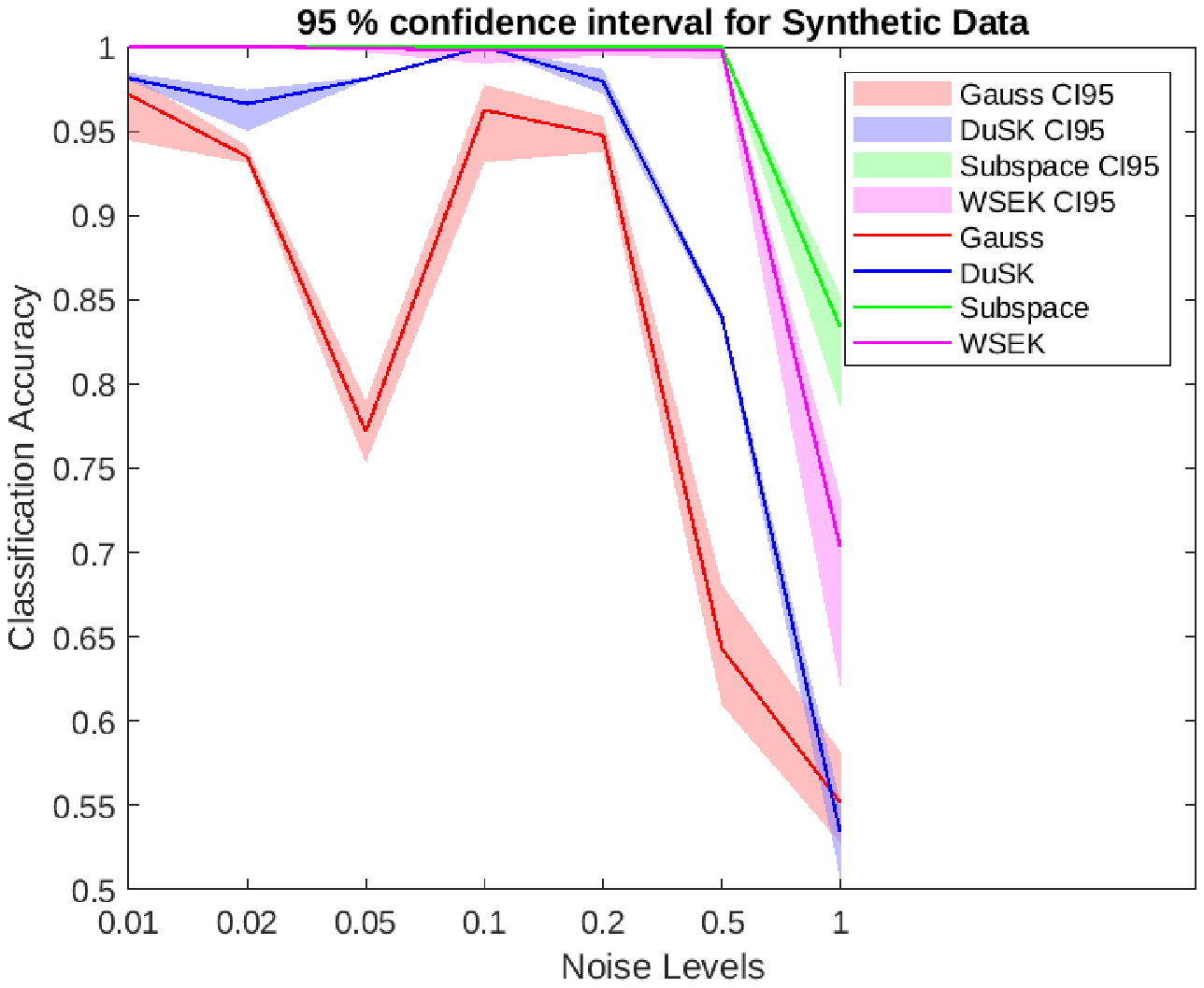}
                	\includegraphics[scale = 0.4]{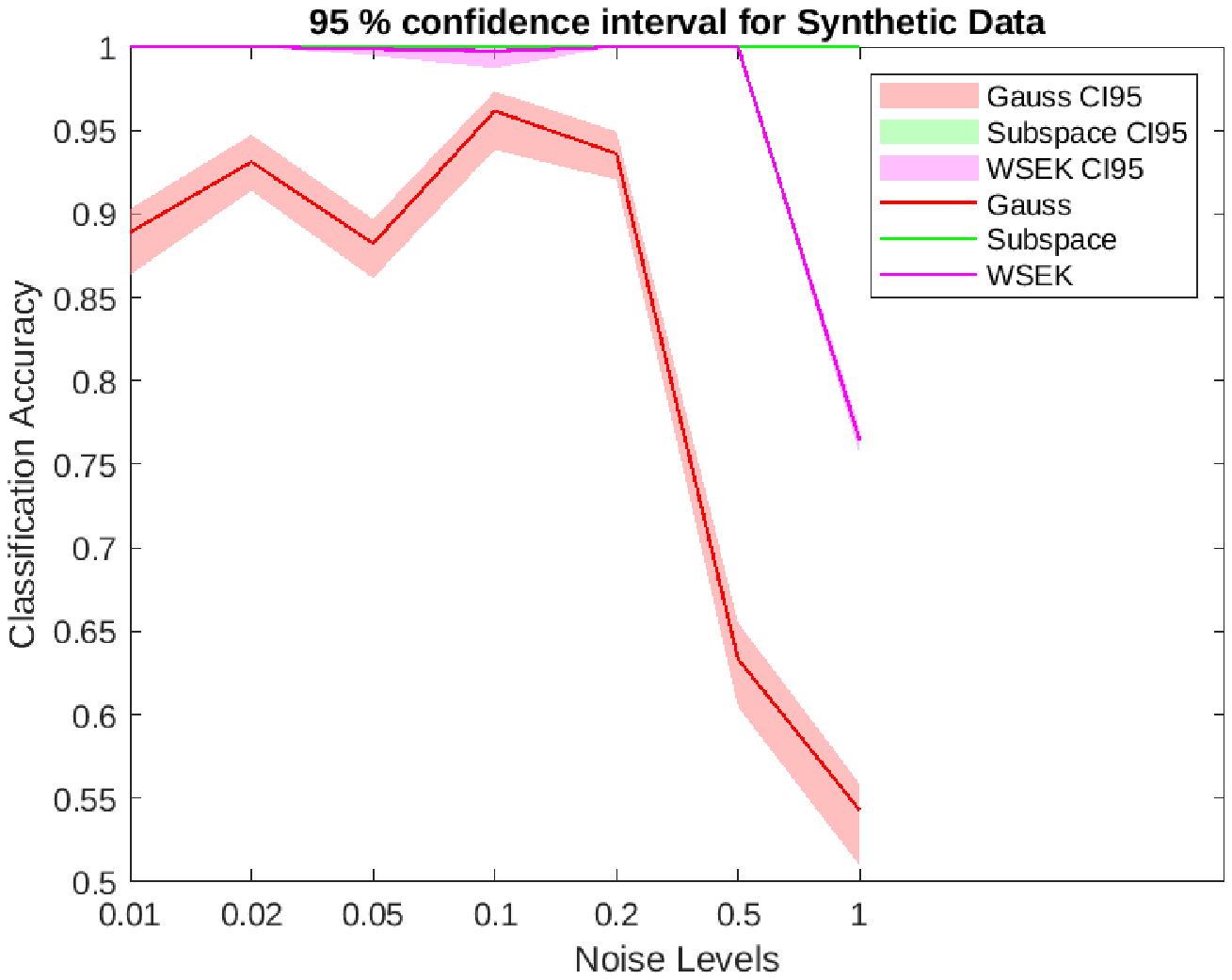}
                	\caption{Classification Accuracy with 95\% confidence interval for different noise levels in the {\em leaf}-scenario of the generated synthetic data with different rank trucation $r_{approx} = 1,3,5,10$ (DuSK not included for $r_{approx} = 10$)}
                \label{fig:synthFreq}
                \end{figure}

                In~Fig.~\ref{fig:synthCore} and Fig.~\ref{fig:synthFreq}, we can see the results for the two experimental scenarios. In both cases, we plotted the test error (i.e.~the classification accuracy on the test set) for the ranks $r_{approx} = 1,3,5,10$. For each rank, we plotted the accuracies of the different kernels for all noise levels in one picture. The computation of DuSK was too expensive in the case $r_{approx} = 10$ as this requires the computation of norms of tensors with CP rank 1000 many times.

                In the {\em core}-scenario, the Gaussian kernel outperforms both DuSK and the Subspace kernel. The WSEK performs similarly to the Gaussian kernel. In the {\em leaf}-scenario, the Subspace kernel gives 100\% accuracy in all cases, and all but the Gaussian kernel performed very well on this data. 

                These results are not surprising following the considerations in Sec.~\ref{sec:intcptuck}: The Subspace kernel sees only the information in the leafs and it can therefore not perform well in the {\em core}-scenario. The DuSK kernel includes extra information so that it performs better in the {\em core}-scenario (especially when we guessed the rank correctly, $r_{approx} = r_{exact} = 3$) but still not as good as the Gaussian kernel. In the {\em leaf}-scenario, all the information is in the subspaces and therefore both DuSK and the Subspace kernel perform very well. Here, the Gaussian kernel performed much worse than all other kernels. The WSEK was designed to do well in both scenarios and this is shown also in these experiments.
                
We will see in the next chapter that especially the {\em leaf}-scenario is realistic: In the ADNI dataset explored below, the subspace kernel performs better than the Gaussian kernel and it even outperforms DuSK. We conclude that including information on the $m$-mode subspaces is crucial for the design of a tensor kernel. This is also a possible explanation why DuSK has performed well in many settings. Our new kernel however outperforms DuSK in all our experiments.

\section{Classification of real datasets}\label{numel}

In this section, we test the performance of the discussed tensor kernels on two real world datasets: The ADNI dataset contains fMRI images of patients with and without Alzheimer's disease and the ADHD dataset contains fMRI images of ADHD patients and healthy subjects. We use the KSTM with the different kernels to distinguish the two classes of subjects in each dataset.

	    All numerical experiments have been done in \texttt{MATLAB 2019b}. Low-rank tensor approximations are computed using \texttt{TT-Toolbox}~\citep{ttTool} and \texttt{tensor toolbox}~\citep{tenTool}. We have run all experiments on a compute cluster which is equipped with \texttt{2 TB NVMe SSD} Harddisk, $2\times$\texttt{Intel Xeon Skylake Silver 4210R} CPUs with 10 cores per CPU, and \texttt{768 GB DDR4 ECC} of RAM.The hyperparameters in KSTM \eqref{eq: kerSTM} are tuned similarly to Synthetic experiments (see Section~\ref{sec:synthexp}), the only difference is that we report results for real data experiments for each of the rank  $R \in \{1,2, \cdots, 10\}$, where $R_1 = R_2 = R_3 = R$. The SVM problem \eqref{eq: kerSTM} is solved using the \texttt{svmtrain} function from LIBSVM~\citep{libsvm} library, and \texttt{svmpredict} computes the classification accuracy using  \eqref{eq:classifier} and \eqref{eq:compbval}.
	    
	                                \begin{figure}
                                \includegraphics[scale = 0.4]{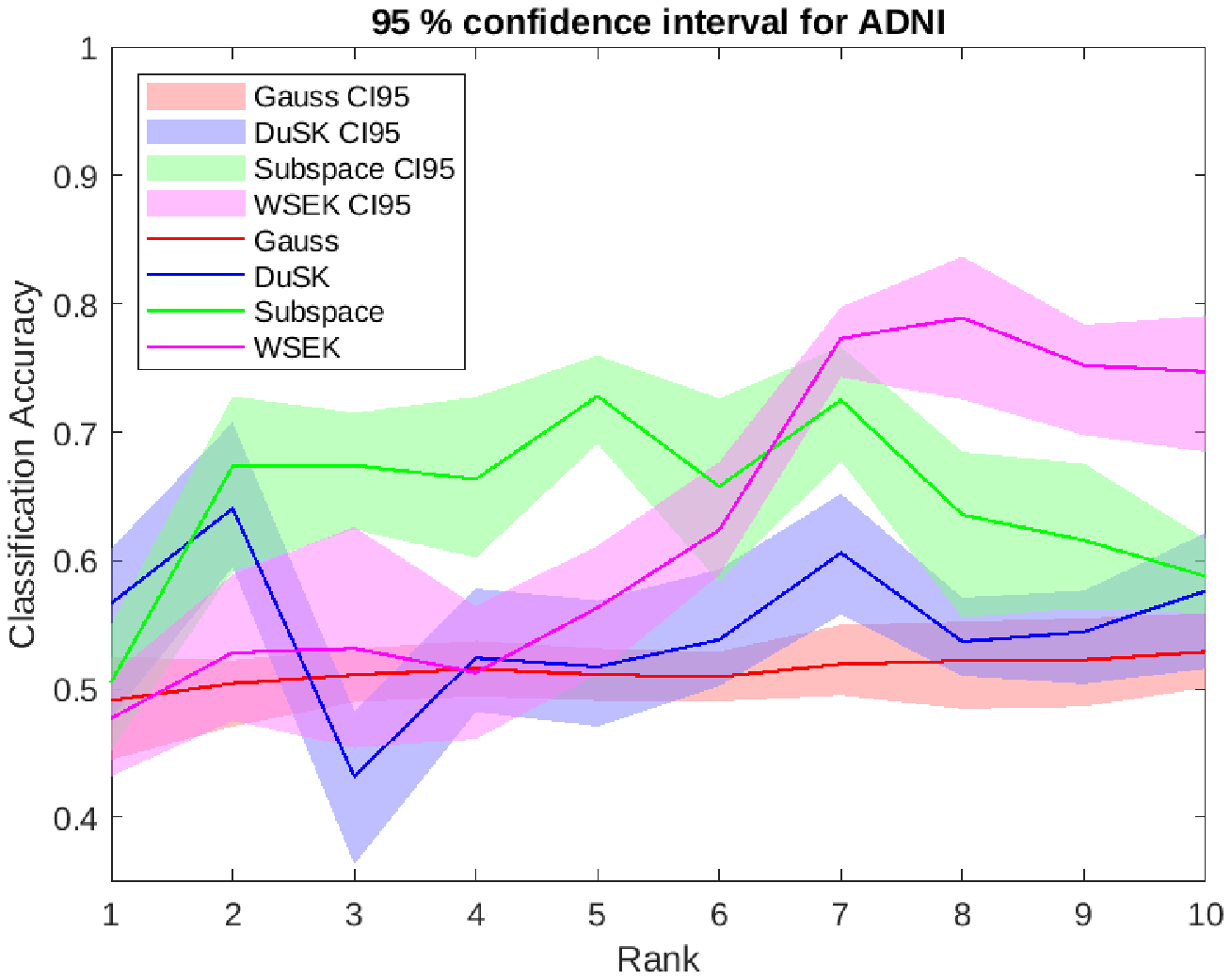}\includegraphics[scale = 0.4]{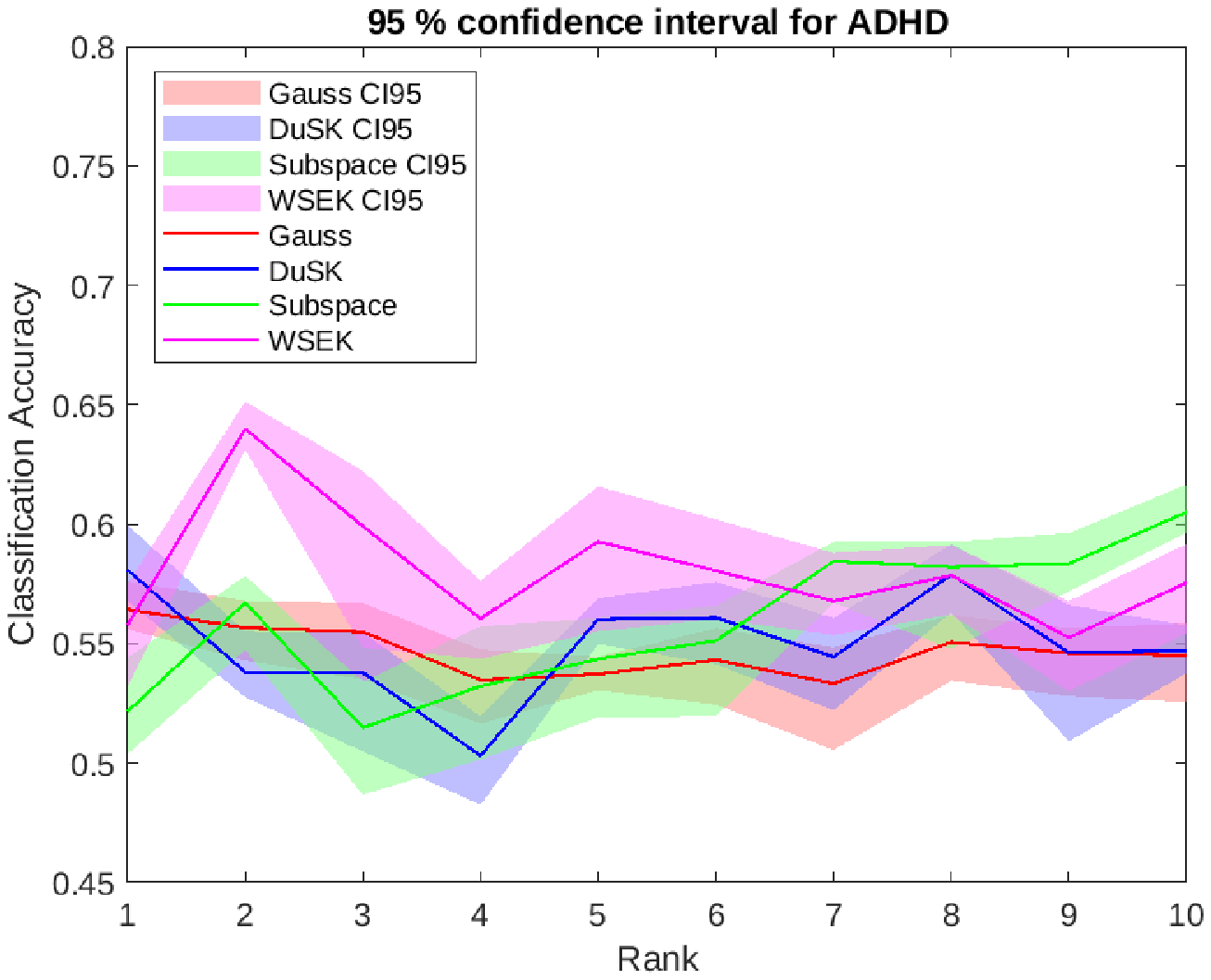}
                                \caption{
                                (left) Comparison of mean classification accuracy with variance for the different kernels with different rank truncation for ADNI dataset, (right) Comparison of mean classification accuracy with variance for the different kernels with different rank truncation for ADHD dataset.}
                                \label{fig:ADNItucker}
                            \end{figure}

                \subsection{Resting-state fMRI data collection}
                    \begin{itemize}
                        \item \textbf{Alzheimer Disease (ADNI)}: The ADNI\footnote{\url{http://adni.loni.usc.edu/}} stands for Alzheimer Disease Neuroimaging Initiative. It contains the resting state fMRI images of 33 subjects. The dataset was collected from the authors of~\cite{MMK}. The images belong to either Mild Cognitive Impairment (MCI) with Alzheimer Disease (AD), or normal controls. Each image is a tensor of size $61 \times 73 \times 61$, containing $271633$ elements in total. The AD+MCI images are labeled with $-1$, and the normal control images are labeled with $1$. Preprocessing of the data sets is explained in~\cite{DuSK}.
                        \item \textbf{Attention Deficit Hyperactivity Disorder (ADHD):} The ADHD dataset is collected from the ADHD-200 global competition dataset\footnote{\url{http://neurobureau.projects.nitrc.org/ADHD200/Data.html}}. It is a publicly available preprocessed fMRI dataset from eight different institutes. The original dataset is unbalanced, so we have chosen 200 subjects randomly, ensuring that 100 of them are ADHD patients (assigned the classification label $-1$) and the 100 other subjects are healthy (denoted with label $1$). Each of the 200 resting state fMRI samples contains $49  \times 58 \times 47  = 133574$ voxels.
		
\begin{remark}
		                The dataset taken here is exactly the same as that used in~\cite{KourJMLR}, so the TT-MMK results can be compared one to one. However, the particular indices of the collected data are not similar to those selected in \cite{DuSK}, so the accuracy of DuSK reported below is not directly comparable to that in \cite{DuSK}.
\end{remark}
		            \end{itemize}

\subsection{Numerical results}

In this section, we summarize the results for the two fMRI datasets:

                            \begin{itemize}
    
                                \item \textbf{Classification accuracy:} In Fig.~\ref{fig:ADNItucker}, we show the average classification accuracy resulting from the cross validation. In Table~\ref{tab:fmriacc}, we show the best classification accuracy between rank $\left[1,10\right]$. On both datasets the proposed WSEK gives the best average classification accuracy
($79\%$ for ADNI and $64\%$ for ADHD) compared to other state of the art tensor kernels. We note that the accuracy of the subspace kernel improves for higher ranks in the ADHD dataset and is then similar to that of WSEK but this choice of rank is very high for a low-rank truncation method. On the other hand, our proposed kernel gives good classification accuracy already at rank 2. The Gaussian kernel was computed for different Tucker approximations of the full tensor. Using the full tensor in the computation of the kernel did not improve the accuracy. As in~\cite{KourJMLR}, the DuSK kernel was computed using a CP approximation of the full tensor (CP-DuSK), because computing the Tucker decomposition and then converting to CP yielded high CP ranks and DuSK was too slow.

                \begin{table}[ht]
                    \caption{Maximum average classification accuracy in percentage $\pm$ standard deviation for different methods, data sets, and rank $R \in \left[1,10\right]$. The values for TTCP-DuSK are taken from~\cite{KourJMLR} for comparison.}
                    \label{tab:fmriacc}
                    \vskip 0.15in
                    \begin{center}
                        \begin{small}
                            \begin{tabular}{lccccr}
                                \hline
                                Methods & ADNI & ADHD   \\
                                \hline
                                Gauss &  53   &  50\\
                                CP-DuSK & 64 $\pm$ 0.05 (R = 5) & 58 $\pm$ 0.02 (R = 6) \\
                                TTCP-DuSK & 73 $\pm$ 0.03 \textbf{(R = 4)} & 63 $\pm$ 0.01 (R = 5)\\
                                Tucker Subspace & $73 \pm 0.04$ (R = 7) & 61 $\pm 0.02$ (R = 10),  \\
                                \textbf{WSEK} & \textbf{79} $\pm 0.03$ (R = 8)  & \textbf{64} $\pm \mathbf{0.01}$ \textbf{(R = 2)}\\
                                \hline
                            \end{tabular}
                        \end{small}
                    \end{center}
                    \vskip -0.1in
                \end{table}

                                \item \textbf{Running Time:} Table~\ref{tab:timecompare} shows the running times for the computation of the different kernels on the ADNI dataset. For small tensors, computing the Gauss kernel is fast. Both the subspace kernel and WSEK can be computed efficiently. Computation of DuSK however quickly becomes prohibitive and takes a long time in these experiments.

 \begin{table}[ht]
                    \caption{Comparison of Kernel computation time for $ R\in \left[ 1,10 \right]$. The values for TTCP-DuSK are taken from~\cite{KourJMLR} for comparison.}
                    \vskip 0.15in
                    \label{tab:timecompare}
                    \begin{center}
                        \begin{tabular}{|c|c|c|c|}
                            \hline 
                           Kernel & Format & Parameters  & CPUtime for ADNI (\# run = 1) \\
                            \hline
                            Gaussian &  Ktensor &  $C, g$   &  $20$ seconds\\
                            \hline
                            DuSK & CP (\ref{subsec:cp})  &  $C, g$,~ $R$   & $17$ minutes \\%\citep{KourJMLR}\\
                            \hline
                            DuSK & TTCP (\ref{sec:convcp})  &   $C, g, ~R_{TT}$    & $ 3.5$ hours \\%\citep{KourJMLR}\\
                            \hline	
                            Subspace & Tucker (\ref{subsec:tucker}) &  $C, g,~ R_{Tucker}$   & $25$ seconds \\
                            \hline
                            WSEK & SqrtmHOSVD (\ref{alg:SqrtmHOSVD}) &  $C, g,~ R_{Tucker}$   & $30$ seconds\\
                            \hline 
                        \end{tabular}
                    \end{center}
                    \vskip 0.15in
                \end{table}
                                
                                \item \textbf{Statistic comparison:} In Tabel~\ref{tab:fmriacc} and Figure~\ref{fig:ADNItucker}, the variance for corresponding mean accuracy is shown. The WSEK STM shows a good trade-off between classification accuracy and variance. For the ADHD dataset, it even gives the best classfication accuracy with the lowest variance value for small ranks.

                            \end{itemize}

\subsection{Conclusion}

Our real world experiments show superior performance of the WSEK both in terms of classification accuracy and running time. We conclude that the classification information of the datasets is hidden mostly in the subspaces of the Tucker decomposition (hence the previously observed good performance of DuSK) but that classification can be improved by taking the singular values into account (as done in WSEK). Furthermore, computing the Tucker decomposition of tensor inputs is straightforward and efficient, resulting in an all-around very robust tensor kernel.

                \section*{Author contributions}
                
                All authors contributed to the conception and design of the tensor kernel method. Implementation and analysis of the new kernel were performed by Kirandeep Kour and Max Pfeffer. The first draft of the manuscript was written by Max Pfeffer and Kirandeep Kour. All authors commented on previous versions of the manuscript. All authors read and approved the final manuscript.
                
                \section*{Acknowledgments}
                K.K. is supported by the International Max Planck Research School for Advanced Methods in Process and System Engineering-\textbf{IMPRS ProEng}, Magdeburg. M.P.~is funded by the Deutsche Forschungsgemeinschaft (DFG, German Research Foundation) – Projektnummer 448293816.
             \subsection*{Statements and Declaration}
             \textbf{Conflict of interest} The authors declare that they have no conflict of interests. Also, there are no financial and non-financial interest to disclose.


\begin{thebibliography}{}
\providecommand{\doi}[1]{\url{https://doi.org/#1}}
\bibcommenthead

\bibitem [\protect \citeauthoryear {%
Acar%
, Kolda%
\BCBL {}\ \BBA {} Dunlavy%
}{%
Acar%
\ \protect \BOthers {.}}{%
{\protect \APACyear {2011}}%
}]{%
Acar2011AllatonceOF}
\APACinsertmetastar {%
Acar2011AllatonceOF}%
\begin{APACrefauthors}%
Acar, E.%
, Kolda, T.G.%
\BCBL {} Dunlavy, D.M.%
\end{APACrefauthors}%
\unskip\
\newblock
\APACrefYearMonthDay{2011}{}{}.
\newblock
{\BBOQ}\APACrefatitle {All-at-once Optimization for Coupled Matrix and Tensor
  Factorizations} {All-at-once optimization for coupled matrix and tensor
  factorizations}.{\BBCQ}
\newblock
\APACjournalVolNumPages{arXiv}{abs/1105.3422}{}{}.
\newblock

\newblock

\PrintBackRefs{\CurrentBib}

\bibitem [\protect \citeauthoryear {%
tensortoolbox Version 3.4}{%
tensortoolbox Version 3.4}{%
{\protect \APACyear {2022}}%
}]{%
tenTool}
\APACinsertmetastar {%
tenTool}%
\begin{APACrefauthors}%
Bader, B.W.%
, Kolda, T.G.%
\BCBL {}\ \BOthersPeriod {.}\end{APACrefauthors}%
\unskip\
\newblock
\APACrefYearMonthDay{2022}{}{}.
\newblock
\APACrefbtitle {Tensor Toolbox for MATLAB Version 3.4.} {Tensor toolbox for
  matlab version 3.4.}
\newblock
\begin{APACrefURL} {https://www.tensortoolbox.org/} \end{APACrefURL}
\PrintBackRefs{\CurrentBib}

\bibitem [\protect \citeauthoryear {%
Cai%
, He%
, Wen%
, Han%
\BCBL {}\ \BBA {} Ma%
}{%
Cai%
\ \protect \BOthers {.}}{%
{\protect \APACyear {2006}}%
}]{%
Cai06supporttensor}
\APACinsertmetastar {%
Cai06supporttensor}%
\begin{APACrefauthors}%
Cai, D.%
, He, X.%
, Wen, J\BHBI R.%
, Han, J.%
\BCBL {} Ma, W\BHBI Y.%
\end{APACrefauthors}%
\unskip\
\newblock
\APACrefYearMonthDay{2006}{}{}.
\newblock
{\BBOQ}\APACrefatitle {Support tensor machines for text categorization}
  {Support tensor machines for text categorization}.{\BBCQ}
\newblock
\APACjournalVolNumPages{The University of Illinois at Urbana-Champaign Computer
  Science Department}{}{}{}.
\newblock

\newblock

\PrintBackRefs{\CurrentBib}

\bibitem [\protect \citeauthoryear {%
Chang%
\ \BBA {} Lin%
}{%
Chang%
\ \BBA {} Lin%
}{%
{\protect \APACyear {2011}}%
}]{%
libsvm}
\APACinsertmetastar {%
libsvm}%
\begin{APACrefauthors}%
Chang, C\BHBI C.%
\BCBT {}\ \BBA {} Lin, C\BHBI J.%
\end{APACrefauthors}%
\unskip\
\newblock
\APACrefYearMonthDay{2011}{May}{}.
\newblock
{\BBOQ}\APACrefatitle {LIBSVM: A Library for Support Vector Machines} {Libsvm:
  A library for support vector machines}.{\BBCQ}
\newblock
\APACjournalVolNumPages{ACM Transactions on Intelligent Systems and
  Technology}{2}{3}{}.
\newblock
\begin{APACrefURL} {https://doi.org/10.1145/1961189.1961199} \end{APACrefURL}
\newblock

\newblock

\PrintBackRefs{\CurrentBib}

\bibitem [\protect \citeauthoryear {%
Chen%
, Batselier%
, Ko%
\BCBL {}\ \BBA {} Wong%
}{%
Chen%
\ \protect \BOthers {.}}{%
{\protect \APACyear {2019}}%
}]{%
chen2018support}
\APACinsertmetastar {%
chen2018support}%
\begin{APACrefauthors}%
Chen, C.%
, Batselier, K.%
, Ko, C\BHBI Y.%
\BCBL {} Wong, N.%
\end{APACrefauthors}%
\unskip\
\newblock
\APACrefYearMonthDay{2019}{}{}.
\newblock
{\BBOQ}\APACrefatitle {A support tensor train machine} {A support tensor train
  machine}.{\BBCQ}
\newblock
 \APACrefbtitle {{2019 International Joint Conference on Neural Networks
  (IJCNN)}} {{2019 International Joint Conference on Neural Networks (IJCNN)}}\
  (\BPGS\ 1--8).
\PrintBackRefs{\CurrentBib}

\bibitem [\protect \citeauthoryear {%
Cichocki%
}{%
Cichocki%
}{%
{\protect \APACyear {2011}}%
}]{%
Cichocki13}
\APACinsertmetastar {%
Cichocki13}%
\begin{APACrefauthors}%
Cichocki, A.%
\end{APACrefauthors}%
\unskip\
\newblock
\APACrefYearMonthDay{2011}{}{}.
\newblock
{\BBOQ}\APACrefatitle {Tensor decompositions: new concepts in brain data
  analysis?} {Tensor decompositions: new concepts in brain data
  analysis?}{\BBCQ}
\newblock
\APACjournalVolNumPages{Journal of the Society of Instrument and Control
  Engineers}{50}{7}{507--516}.
\newblock

\newblock

\PrintBackRefs{\CurrentBib}

\bibitem [\protect \citeauthoryear {%
Cichocki%
\ \protect \BOthers {.}}{%
Cichocki%
\ \protect \BOthers {.}}{%
{\protect \APACyear {2016}}%
}]{%
cichocki2016tensor}
\APACinsertmetastar {%
cichocki2016tensor}%
\begin{APACrefauthors}%
Cichocki, A.%
, Lee, N.%
, Oseledets, I.%
, Phan, A\BHBI H.%
, Zhao, Q.%
\BCBL {} Mandic, D.P.%
\end{APACrefauthors}%
\unskip\
\newblock
\APACrefYearMonthDay{2016}{}{}.
\newblock
{\BBOQ}\APACrefatitle {Tensor Networks for Dimensionality Reduction and
  Large-scale Optimization: Part 1 Low-Rank Tensor Decompositions} {Tensor
  networks for dimensionality reduction and large-scale optimization: Part 1
  low-rank tensor decompositions}.{\BBCQ}
\newblock
\APACjournalVolNumPages{FNT in Machine Learning}{9}{4-5}{249-429}.
\newblock

\newblock

\PrintBackRefs{\CurrentBib}

\bibitem [\protect \citeauthoryear {%
Cortes%
\ \BBA {} Vapnik%
}{%
Cortes%
\ \BBA {} Vapnik%
}{%
{\protect \APACyear {1995}}%
}]{%
vapnik95}
\APACinsertmetastar {%
vapnik95}%
\begin{APACrefauthors}%
Cortes, C.%
\BCBT {}\ \BBA {} Vapnik, V.%
\end{APACrefauthors}%
\unskip\
\newblock
\APACrefYearMonthDay{1995}{}{}.
\newblock
{\BBOQ}\APACrefatitle {Support-vector networks} {Support-vector
  networks}.{\BBCQ}
\newblock
\APACjournalVolNumPages{Machine Learning}{20}{}{273-297}.
\newblock

\newblock

\PrintBackRefs{\CurrentBib}

\bibitem [\protect \citeauthoryear {%
de Silva%
\ \BBA {} Lim%
}{%
de Silva%
\ \BBA {} Lim%
}{%
{\protect \APACyear {2008}}%
}]{%
desilva2008}
\APACinsertmetastar {%
desilva2008}%
\begin{APACrefauthors}%
de Silva, V.%
\BCBT {}\ \BBA {} Lim, L\BHBI H.%
\end{APACrefauthors}%
\unskip\
\newblock
\APACrefYearMonthDay{2008}{}{}.
\newblock
{\BBOQ}\APACrefatitle {Tensor rank and the ill-posedness of the best low-rank
  approximation problem} {Tensor rank and the ill-posedness of the best
  low-rank approximation problem}.{\BBCQ}
\newblock
\APACjournalVolNumPages{SIAM Journal on Matrix Analysis and
  Applications}{30}{3}{1084--1127}.
\newblock

\newblock

\PrintBackRefs{\CurrentBib}

\bibitem [\protect \citeauthoryear {%
Guo%
, Kotsia%
\BCBL {}\ \BBA {} Patras%
}{%
Guo%
\ \protect \BOthers {.}}{%
{\protect \APACyear {2012}}%
}]{%
Guo12}
\APACinsertmetastar {%
Guo12}%
\begin{APACrefauthors}%
Guo, W.%
, Kotsia, I.%
\BCBL {} Patras, I.%
\end{APACrefauthors}%
\unskip\
\newblock
\APACrefYearMonthDay{2012}{}{}.
\newblock
{\BBOQ}\APACrefatitle {Tensor Learning for Regression} {Tensor learning for
  regression}.{\BBCQ}
\newblock
\APACjournalVolNumPages{IEEE Transactions on Image Processing}{21}{2}{816-827}.
\newblock

\newblock

\PrintBackRefs{\CurrentBib}

\bibitem [\protect \citeauthoryear {%
Hao%
, He%
, Chen%
\BCBL {}\ \BBA {} Yang%
}{%
Hao%
\ \protect \BOthers {.}}{%
{\protect \APACyear {2013}}%
}]{%
Hao13}
\APACinsertmetastar {%
Hao13}%
\begin{APACrefauthors}%
Hao, Z.%
, He, L.%
, Chen, B.%
\BCBL {} Yang, X.%
\end{APACrefauthors}%
\unskip\
\newblock
\APACrefYearMonthDay{2013}{}{}.
\newblock
{\BBOQ}\APACrefatitle {{A} Linear Support Higher-Order Tensor Machine for
  Classification} {{A} linear support higher-order tensor machine for
  classification}.{\BBCQ}
\newblock
\APACjournalVolNumPages{IEEE Transactions on Image
  Processing}{22}{7}{2911-2920}.
\newblock

\newblock

\PrintBackRefs{\CurrentBib}

\bibitem [\protect \citeauthoryear {%
He%
\ \protect \BOthers {.}}{%
He%
\ \protect \BOthers {.}}{%
{\protect \APACyear {2014}}%
}]{%
DuSK}
\APACinsertmetastar {%
DuSK}%
\begin{APACrefauthors}%
He, L.%
, Kong, X.%
, Yu, P.S.%
, Yang, X.%
, Ragin, A.B.%
\BCBL {} Hao, Z.%
\end{APACrefauthors}%
\unskip\
\newblock
\APACrefYearMonthDay{2014}{}{}.
\newblock
{\BBOQ}\APACrefatitle {DuSK: A Dual Structure-preserving Kernel for Supervised
  Tensor Learning with Applications to Neuroimages} {Dusk: A dual
  structure-preserving kernel for supervised tensor learning with applications
  to neuroimages}.{\BBCQ}
\newblock
\APACjournalVolNumPages{Proceedings of the 2014 SIAM International Conference
  on Data Mining (SDM)}{}{}{127-135}.
\newblock
\begin{APACrefURL} {https://epubs.siam.org/doi/abs/10.1137/1.9781611973440.15}
  \end{APACrefURL}
\newblock

\newblock

\PrintBackRefs{\CurrentBib}

\bibitem [\protect \citeauthoryear {%
He%
, Lu%
, Ding%
\BCBL {}\ \protect \BOthers {.}}{%
He%
, Lu%
, Ding%
\BCBL {}\ \protect \BOthers {.}}{%
{\protect \APACyear {2017}}%
}]{%
MMK}
\APACinsertmetastar {%
MMK}%
\begin{APACrefauthors}%
He, L.%
, Lu, C\BHBI T.%
, Ding, H.%
, Wang, S.%
, Shen, L.%
, Yu, P.S.%
\BCBL {} Ragin, A.B.%
\end{APACrefauthors}%
\unskip\
\newblock
\APACrefYearMonthDay{2017}{}{}.
\newblock
{\BBOQ}\APACrefatitle {Multi-Way Multi-Level Kernel Modeling for Neuroimaging
  Classification} {Multi-way multi-level kernel modeling for neuroimaging
  classification}.{\BBCQ}
\newblock
\APACjournalVolNumPages{The IEEE Conference on Computer Vision and Pattern
  Recognition (CVPR)}{}{}{6846-6854}.
\newblock

\newblock

\PrintBackRefs{\CurrentBib}

\bibitem [\protect \citeauthoryear {%
He%
, Lu%
, Ma%
\BCBL {}\ \protect \BOthers {.}}{%
He%
, Lu%
, Ma%
\BCBL {}\ \protect \BOthers {.}}{%
{\protect \APACyear {2017}}%
}]{%
KSTM17}
\APACinsertmetastar {%
KSTM17}%
\begin{APACrefauthors}%
He, L.%
, Lu, C\BHBI T.%
, Ma, G.%
, Wang, S.%
, Shen, L.%
, Yu, P.S.%
\BCBL {} Ragin, A.B.%
\end{APACrefauthors}%
\unskip\
\newblock
\APACrefYearMonthDay{2017}{}{}.
\newblock
{\BBOQ}\APACrefatitle {Kernelized support tensor machines} {Kernelized support
  tensor machines}.{\BBCQ}
\newblock
\APACjournalVolNumPages{Proceedings of the 34th International Conference on
  Machine Learning-Volume 70}{}{}{1442--1451}.
\newblock

\newblock

\PrintBackRefs{\CurrentBib}

\bibitem [\protect \citeauthoryear {%
Hitchcock%
}{%
Hitchcock%
}{%
{\protect \APACyear {1927}}%
}]{%
Hitchcock1927}
\APACinsertmetastar {%
Hitchcock1927}%
\begin{APACrefauthors}%
Hitchcock, F.L.%
\end{APACrefauthors}%
\unskip\
\newblock
\APACrefYearMonthDay{1927}{}{}.
\newblock
{\BBOQ}\APACrefatitle {The Expression of a Tensor or a Polyadic as a Sum of
  Products} {The expression of a tensor or a polyadic as a sum of
  products}.{\BBCQ}
\newblock
\APACjournalVolNumPages{Journal of Mathematics and Physics}{6}{1-4}{164-189}.
\newblock

\newblock

\PrintBackRefs{\CurrentBib}

\bibitem [\protect \citeauthoryear {%
Kolda%
\ \BBA {} Bader%
}{%
Kolda%
\ \BBA {} Bader%
}{%
{\protect \APACyear {2009}}%
}]{%
Kolda09}
\APACinsertmetastar {%
Kolda09}%
\begin{APACrefauthors}%
Kolda, T.G.%
\BCBT {}\ \BBA {} Bader, B.W.%
\end{APACrefauthors}%
\unskip\
\newblock
\APACrefYearMonthDay{2009}{}{}.
\newblock
{\BBOQ}\APACrefatitle {Tensor Decompositions and Applications} {Tensor
  decompositions and applications}.{\BBCQ}
\newblock
\APACjournalVolNumPages{SIAM Review}{51}{3}{455--500}.
\newblock

\newblock

\PrintBackRefs{\CurrentBib}

\bibitem [\protect \citeauthoryear {%
Kotsia%
\ \BBA {} Patras%
}{%
Kotsia%
\ \BBA {} Patras%
}{%
{\protect \APACyear {2011}}%
}]{%
kotsia}
\APACinsertmetastar {%
kotsia}%
\begin{APACrefauthors}%
Kotsia, I.%
\BCBT {}\ \BBA {} Patras, I.%
\end{APACrefauthors}%
\unskip\
\newblock
\APACrefYearMonthDay{2011}{June}{}.
\newblock
{\BBOQ}\APACrefatitle {{Support Tucker Machines}} {{Support Tucker
  Machines}}.{\BBCQ}
\newblock
\APACjournalVolNumPages{CVPR 2011}{}{}{633-640}.
\newblock

\newblock

\PrintBackRefs{\CurrentBib}

\bibitem [\protect \citeauthoryear {%
Kour%
, Dolgov%
, Stoll%
\BCBL {}\ \BBA {} Benner%
}{%
Kour%
\ \protect \BOthers {.}}{%
{\protect \APACyear {2023}}%
}]{%
KourJMLR}
\APACinsertmetastar {%
KourJMLR}%
\begin{APACrefauthors}%
Kour, K.%
, Dolgov, S.%
, Stoll, M.%
\BCBL {} Benner, P.%
\end{APACrefauthors}%
\unskip\
\newblock
\APACrefYearMonthDay{2023}{}{}.
\newblock
{\BBOQ}\APACrefatitle {Efficient Structure-preserving Support Tensor Train
  Machine} {Efficient structure-preserving support tensor train
  machine}.{\BBCQ}
\newblock
\APACjournalVolNumPages{Journal of Machine Learning Research}{24}{4}{1--22}.
\newblock
\begin{APACrefURL} {http://jmlr.org/papers/v24/20-1310.html} \end{APACrefURL}
\newblock

\newblock

\PrintBackRefs{\CurrentBib}

\bibitem [\protect \citeauthoryear {%
Lathauwer%
, Moor%
\BCBL {}\ \BBA {} Vandewalle%
}{%
Lathauwer%
\ \protect \BOthers {.}}{%
{\protect \APACyear {2000}}%
}]{%
TuckerLathauwer}
\APACinsertmetastar {%
TuckerLathauwer}%
\begin{APACrefauthors}%
Lathauwer, L.D.%
, Moor, B.D.%
\BCBL {} Vandewalle, J.%
\end{APACrefauthors}%
\unskip\
\newblock
\APACrefYearMonthDay{2000}{}{}.
\newblock
{\BBOQ}\APACrefatitle {A Multilinear Singular Value Decomposition} {A
  multilinear singular value decomposition}.{\BBCQ}
\newblock
\APACjournalVolNumPages{SIAM Journal on Matrix Analysis and
  Applications}{21}{4}{1253-1278}.
\newblock

\newblock

\PrintBackRefs{\CurrentBib}

\bibitem [\protect \citeauthoryear {%
Liu%
, Guo%
, He%
\BCBL {}\ \BBA {} Yang%
}{%
Liu%
\ \protect \BOthers {.}}{%
{\protect \APACyear {2015}}%
}]{%
Liu15}
\APACinsertmetastar {%
Liu15}%
\begin{APACrefauthors}%
Liu, X.%
, Guo, T.%
, He, L.%
\BCBL {} Yang, X.%
\end{APACrefauthors}%
\unskip\
\newblock
\APACrefYearMonthDay{2015}{}{}.
\newblock
{\BBOQ}\APACrefatitle {A Low-Rank Approximation-Based Transductive Support
  Tensor Machine for Semisupervised Classification} {A low-rank
  approximation-based transductive support tensor machine for semisupervised
  classification}.{\BBCQ}
\newblock
\APACjournalVolNumPages{IEEE Transactions on Image
  Processing}{24}{6}{1825-1838}.
\newblock

\newblock

\PrintBackRefs{\CurrentBib}

\bibitem [\protect \citeauthoryear {%
Nion%
\ \BBA {} Lathauwer%
}{%
Nion%
\ \BBA {} Lathauwer%
}{%
{\protect \APACyear {2008}}%
}]{%
cp_als}
\APACinsertmetastar {%
cp_als}%
\begin{APACrefauthors}%
Nion, D.%
\BCBT {}\ \BBA {} Lathauwer, L.D.%
\end{APACrefauthors}%
\unskip\
\newblock
\APACrefYearMonthDay{2008}{}{}.
\newblock
{\BBOQ}\APACrefatitle {Fast Communication: An Enhanced Line Search Scheme for
  Complex-valued Tensor Decompositions} {Fast communication: An enhanced line
  search scheme for complex-valued tensor decompositions}.{\BBCQ}
\newblock
\APACjournalVolNumPages{Signal Processing}{88}{3}{749--755}.
\newblock

\newblock

\PrintBackRefs{\CurrentBib}

\bibitem [\protect \citeauthoryear {%
TT-Toolbox}{%
TT-Toolbox}{%
{\protect \APACyear {2023}}%
}]{%
ttTool}
\APACinsertmetastar {%
ttTool}%
\begin{APACrefauthors}%
Oseledets, I.%
\BCBT {}\ \BBA {} Dolgov, S.%
\end{APACrefauthors}%
\unskip\
\newblock
\APACrefYearMonthDay{2023}{}{}.
\newblock
\APACrefbtitle {TT-Toolbox.} {Tt-toolbox.}
\newblock
\APACaddressPublisher{}{GitHub}.
\newblock
\begin{APACrefURL} {https://github.com/oseledets/TT-Toolbox} \end{APACrefURL}
\PrintBackRefs{\CurrentBib}

\bibitem [\protect \citeauthoryear {%
I.~Oseledets%
\ \BBA {} Tyrtyshnikov%
}{%
I.~Oseledets%
\ \BBA {} Tyrtyshnikov%
}{%
{\protect \APACyear {2010}}%
}]{%
ttcross}
\APACinsertmetastar {%
ttcross}%
\begin{APACrefauthors}%
Oseledets, I.%
\BCBT {}\ \BBA {} Tyrtyshnikov, E.%
\end{APACrefauthors}%
\unskip\
\newblock
\APACrefYearMonthDay{2010}{}{}.
\newblock
{\BBOQ}\APACrefatitle {TT-cross approximation for multidimensional arrays}
  {Tt-cross approximation for multidimensional arrays}.{\BBCQ}
\newblock
\APACjournalVolNumPages{Linear Algebra and its Applications}{432}{1}{70-88}.
\newblock
\begin{APACrefURL}
  {https://www.sciencedirect.com/science/article/pii/S0024379509003747}
  \end{APACrefURL}
\newblock

\newblock

\PrintBackRefs{\CurrentBib}

\bibitem [\protect \citeauthoryear {%
I.V.~Oseledets%
}{%
I.V.~Oseledets%
}{%
{\protect \APACyear {2011}}%
}]{%
oseledets2011tensor}
\APACinsertmetastar {%
oseledets2011tensor}%
\begin{APACrefauthors}%
Oseledets, I.V.%
\end{APACrefauthors}%
\unskip\
\newblock
\APACrefYearMonthDay{2011}{}{}.
\newblock
{\BBOQ}\APACrefatitle {Tensor-train decomposition} {Tensor-train
  decomposition}.{\BBCQ}
\newblock
\APACjournalVolNumPages{SIAM Journal on Scientific
  Computing}{33}{5}{2295--2317}.
\newblock

\newblock

\PrintBackRefs{\CurrentBib}

\bibitem [\protect \citeauthoryear {%
Pirsiavash%
, Ramanan%
\BCBL {}\ \BBA {} Fowlkes%
}{%
Pirsiavash%
\ \protect \BOthers {.}}{%
{\protect \APACyear {2009}}%
}]{%
Pirisiavash}
\APACinsertmetastar {%
Pirisiavash}%
\begin{APACrefauthors}%
Pirsiavash, H.%
, Ramanan, D.%
\BCBL {} Fowlkes, C.C.%
\end{APACrefauthors}%
\unskip\
\newblock
\APACrefYearMonthDay{2009}{}{}.
\newblock
{\BBOQ}\APACrefatitle {Bilinear classifiers for visual recognition} {Bilinear
  classifiers for visual recognition}.{\BBCQ}
\newblock
\APACjournalVolNumPages{Advances in Neural Information Processing Systems
  22}{}{}{1482--1490}.
\newblock

\newblock

\PrintBackRefs{\CurrentBib}

\bibitem [\protect \citeauthoryear {%
Signoretto%
, Olivetti%
, Lathauwer%
\BCBL {}\ \BBA {} Suykens%
}{%
Signoretto%
\ \protect \BOthers {.}}{%
{\protect \APACyear {2011}}%
}]{%
Signoretto11}
\APACinsertmetastar {%
Signoretto11}%
\begin{APACrefauthors}%
Signoretto, M.%
, Olivetti, E.%
, Lathauwer, L.D.%
\BCBL {} Suykens, J.A.K.%
\end{APACrefauthors}%
\unskip\
\newblock
\APACrefYearMonthDay{2011}{}{}.
\newblock
{\BBOQ}\APACrefatitle {A kernel-based framework to tensorial data analysis} {A
  kernel-based framework to tensorial data analysis}.{\BBCQ}
\newblock
\APACjournalVolNumPages{Neural Networks}{24}{8}{861 - 874}.
\newblock

\newblock

\PrintBackRefs{\CurrentBib}

\bibitem [\protect \citeauthoryear {%
Signoretto%
, Olivetti%
, Lathauwer%
\BCBL {}\ \BBA {} Suykens%
}{%
Signoretto%
\ \protect \BOthers {.}}{%
{\protect \APACyear {2012}}%
}]{%
Signoretto12}
\APACinsertmetastar {%
Signoretto12}%
\begin{APACrefauthors}%
Signoretto, M.%
, Olivetti, E.%
, Lathauwer, L.D.%
\BCBL {} Suykens, J.A.K.%
\end{APACrefauthors}%
\unskip\
\newblock
\APACrefYearMonthDay{2012}{}{}.
\newblock
{\BBOQ}\APACrefatitle {Classification of Multichannel Signals With
  Cumulant-Based Kernels} {Classification of multichannel signals with
  cumulant-based kernels}.{\BBCQ}
\newblock
\APACjournalVolNumPages{IEEE Transactions on Signal
  Processing}{60}{5}{2304-2314}.
\newblock

\newblock

\PrintBackRefs{\CurrentBib}

\bibitem [\protect \citeauthoryear {%
Signoretto%
, Tran~Dinh%
, De~Lathauwer%
\BCBL {}\ \BBA {} Suykens%
}{%
Signoretto%
\ \protect \BOthers {.}}{%
{\protect \APACyear {2014}}%
}]{%
Signoretto2013}
\APACinsertmetastar {%
Signoretto2013}%
\begin{APACrefauthors}%
Signoretto, M.%
, Tran~Dinh, Q.%
, De~Lathauwer, L.%
\BCBL {} Suykens, J.A.K.%
\end{APACrefauthors}%
\unskip\
\newblock
\APACrefYearMonthDay{2014}{}{}.
\newblock
{\BBOQ}\APACrefatitle {Learning with tensors: a framework based on convex
  optimization and spectral regularization} {Learning with tensors: a framework
  based on convex optimization and spectral regularization}.{\BBCQ}
\newblock
\APACjournalVolNumPages{Machine Learning}{94}{3}{303--351}.
\newblock
\begin{APACrefURL} {https://doi.org/10.1007/s10994-013-5366-3} \end{APACrefURL}
\newblock

\newblock

\PrintBackRefs{\CurrentBib}

\bibitem [\protect \citeauthoryear {%
Taguchi%
\ \BBA {} Turki%
}{%
Taguchi%
\ \BBA {} Turki%
}{%
{\protect \APACyear {2021}}%
}]{%
Taguchi2021}
\APACinsertmetastar {%
Taguchi2021}%
\begin{APACrefauthors}%
Taguchi, Y\BHBI H.%
\BCBT {}\ \BBA {} Turki, T.%
\end{APACrefauthors}%
\unskip\
\newblock
\APACrefYearMonthDay{2021}{}{}.
\newblock
{\BBOQ}\APACrefatitle {Mathematical formulation and application of kernel
  tensor decomposition based unsupervised feature extraction} {Mathematical
  formulation and application of kernel tensor decomposition based unsupervised
  feature extraction}.{\BBCQ}
\newblock
\APACjournalVolNumPages{Knowledge-Based Systems}{217}{}{106834}.
\newblock
\begin{APACrefURL} {https://doi.org/10.1016/j.knosys.2021.106834}
  \end{APACrefURL}
\newblock

\newblock

\PrintBackRefs{\CurrentBib}

\bibitem [\protect \citeauthoryear {%
Tao%
, Li%
, Hu%
, Maybank%
\BCBL {}\ \BBA {} Wu%
}{%
Tao%
\ \protect \BOthers {.}}{%
{\protect \APACyear {2005}}%
}]{%
Tao05}
\APACinsertmetastar {%
Tao05}%
\begin{APACrefauthors}%
Tao, D.%
, Li, X.%
, Hu, W.%
, Maybank, S.%
\BCBL {} Wu, X.%
\end{APACrefauthors}%
\unskip\
\newblock
\APACrefYearMonthDay{2005}{}{}.
\newblock
{\BBOQ}\APACrefatitle {Supervised Tensor Learning} {Supervised tensor
  learning}.{\BBCQ}
\newblock
\APACjournalVolNumPages{Proceedings of the Fifth IEEE International Conference
  on Data Mining}{}{}{450-457}.
\newblock
\begin{APACrefURL} {https://doi.org/10.1109/ICDM.2005.139} \end{APACrefURL}
\newblock

\newblock

\PrintBackRefs{\CurrentBib}

\bibitem [\protect \citeauthoryear {%
Tao%
, Li%
, Hu%
, Maybank%
\BCBL {}\ \BBA {} Wu%
}{%
Tao%
\ \protect \BOthers {.}}{%
{\protect \APACyear {2007}}%
}]{%
Tao07}
\APACinsertmetastar {%
Tao07}%
\begin{APACrefauthors}%
Tao, D.%
, Li, X.%
, Hu, W.%
, Maybank, S.%
\BCBL {} Wu, X.%
\end{APACrefauthors}%
\unskip\
\newblock
\APACrefYearMonthDay{2007}{}{}.
\newblock
{\BBOQ}\APACrefatitle {Supervised tensor learning} {Supervised tensor
  learning}.{\BBCQ}
\newblock
\APACjournalVolNumPages{Knowledge and Information Systems}{}{}{1-42}.
\newblock

\newblock

\PrintBackRefs{\CurrentBib}

\bibitem [\protect \citeauthoryear {%
Tucker%
}{%
Tucker%
}{%
{\protect \APACyear {1966}}%
}]{%
tucker1966}
\APACinsertmetastar {%
tucker1966}%
\begin{APACrefauthors}%
Tucker, L.R.%
\end{APACrefauthors}%
\unskip\
\newblock
\APACrefYearMonthDay{1966}{}{}.
\newblock
{\BBOQ}\APACrefatitle {Some mathematical notes on three-mode factor analysis}
  {Some mathematical notes on three-mode factor analysis}.{\BBCQ}
\newblock
\APACjournalVolNumPages{Psychometrika}{31}{}{279-311}.
\newblock
\begin{APACrefURL} {https://doi.org/10.1007/BF02289464} \end{APACrefURL}
\newblock

\newblock

\PrintBackRefs{\CurrentBib}

\bibitem [\protect \citeauthoryear {%
Vannieuwenhoven%
, Vandebril%
\BCBL {}\ \BBA {} Meerbergen%
}{%
Vannieuwenhoven%
\ \protect \BOthers {.}}{%
{\protect \APACyear {2012}}%
}]{%
st_hosvd}
\APACinsertmetastar {%
st_hosvd}%
\begin{APACrefauthors}%
Vannieuwenhoven, N.%
, Vandebril, R.%
\BCBL {} Meerbergen, K.%
\end{APACrefauthors}%
\unskip\
\newblock
\APACrefYearMonthDay{2012}{}{}.
\newblock
{\BBOQ}\APACrefatitle {A New Truncation Strategy for the Higher-Order Singular
  Value Decomposition} {A new truncation strategy for the higher-order singular
  value decomposition}.{\BBCQ}
\newblock
\APACjournalVolNumPages{SIAM Journal on Scientific
  Computing}{34}{2}{A1027-A1052}.
\newblock

\newblock

\PrintBackRefs{\CurrentBib}

\bibitem [\protect \citeauthoryear {%
Vapnik%
}{%
Vapnik%
}{%
{\protect \APACyear {1995}}%
}]{%
vapnik}
\APACinsertmetastar {%
vapnik}%
\begin{APACrefauthors}%
Vapnik, V.%
\end{APACrefauthors}%
\unskip\
\newblock
\APACrefYear{1995}.
\newblock
\APACrefbtitle {The nature of statistical learning theory} {The nature of
  statistical learning theory}.
\newblock
\APACaddressPublisher{New York}{Springer-Verlag}.
\PrintBackRefs{\CurrentBib}

\bibitem [\protect \citeauthoryear {%
Vapnik%
}{%
Vapnik%
}{%
{\protect \APACyear {1998}}%
}]{%
vapnik98}
\APACinsertmetastar {%
vapnik98}%
\begin{APACrefauthors}%
Vapnik, V.%
\end{APACrefauthors}%
\unskip\
\newblock
\APACrefYear{1998}.
\newblock
\APACrefbtitle {Statistical Learning Theory} {Statistical learning theory}.
\newblock
\APACaddressPublisher{}{Wiley-Interscience}.
\PrintBackRefs{\CurrentBib}

\bibitem [\protect \citeauthoryear {%
Wolf%
, Jhuang%
\BCBL {}\ \BBA {} Hazan%
}{%
Wolf%
\ \protect \BOthers {.}}{%
{\protect \APACyear {2007}}%
}]{%
wolf}
\APACinsertmetastar {%
wolf}%
\begin{APACrefauthors}%
Wolf, L.%
, Jhuang, H.%
\BCBL {} Hazan, T.%
\end{APACrefauthors}%
\unskip\
\newblock
\APACrefYearMonthDay{2007}{}{}.
\newblock
{\BBOQ}\APACrefatitle {Modeling Appearances with Low-Rank {SVM}} {Modeling
  appearances with low-rank {SVM}}.{\BBCQ}
\newblock
\APACjournalVolNumPages{IEEE Conference on Computer Vision and Pattern
  Recognition}{}{}{}.
\newblock

\newblock

\PrintBackRefs{\CurrentBib}

\bibitem [\protect \citeauthoryear {%
Yan%
\ \protect \BOthers {.}}{%
Yan%
\ \protect \BOthers {.}}{%
{\protect \APACyear {2007}}%
}]{%
FaceReco2007}
\APACinsertmetastar {%
FaceReco2007}%
\begin{APACrefauthors}%
Yan, S.%
, Xu, D.%
, Yang, Q.%
, Zhang, L.%
, Tang, X.%
\BCBL {} Zhang, H\BHBI J.%
\end{APACrefauthors}%
\unskip\
\newblock
\APACrefYearMonthDay{2007}{}{}.
\newblock
{\BBOQ}\APACrefatitle {Multilinear Discriminant Analysis for Face Recognition}
  {Multilinear discriminant analysis for face recognition}.{\BBCQ}
\newblock
\APACjournalVolNumPages{IEEE Transactions on Image Processing}{16}{1}{212-220}.
\newblock

\newblock

\PrintBackRefs{\CurrentBib}

\bibitem [\protect \citeauthoryear {%
Zeng%
, Wang%
, Shen%
\BCBL {}\ \BBA {} Shi%
}{%
Zeng%
\ \protect \BOthers {.}}{%
{\protect \APACyear {2017}}%
}]{%
Zeng17}
\APACinsertmetastar {%
Zeng17}%
\begin{APACrefauthors}%
Zeng, D.%
, Wang, S.%
, Shen, Y.%
\BCBL {} Shi, C.%
\end{APACrefauthors}%
\unskip\
\newblock
\APACrefYearMonthDay{2017}{}{}.
\newblock
{\BBOQ}\APACrefatitle {A {GA}-based feature selection and parameter
  optimization for support tucker machine} {A {GA}-based feature selection and
  parameter optimization for support tucker machine}.{\BBCQ}
\newblock
\APACjournalVolNumPages{Procedia Computer Science}{111}{}{17 - 23}.
\newblock

\newblock

\PrintBackRefs{\CurrentBib}

\bibitem [\protect \citeauthoryear {%
Zhao%
, Zhou%
, Adali%
, Zhang%
\BCBL {}\ \BBA {} Cichocki%
}{%
Zhao%
\ \protect \BOthers {.}}{%
{\protect \APACyear {2013}}%
{\protect \APACexlab {{\protect \BCnt {1}}}}}]{%
QZhao13a}
\APACinsertmetastar {%
QZhao13a}%
\begin{APACrefauthors}%
Zhao, Q.%
, Zhou, G.%
, Adali, T.%
, Zhang, L.%
\BCBL {} Cichocki, A.%
\end{APACrefauthors}%
\unskip\
\newblock
\APACrefYearMonthDay{2013{\protect \BCnt {1}}}{}{}.
\newblock
{\BBOQ}\APACrefatitle {Kernelization of Tensor-Based Models for Multiway Data
  Analysis: Processing of Multidimensional Structured Data} {Kernelization of
  tensor-based models for multiway data analysis: Processing of
  multidimensional structured data}.{\BBCQ}
\newblock
\APACjournalVolNumPages{IEEE Signal Processing Magazine}{30}{4}{137-148}.
\newblock

\newblock

\PrintBackRefs{\CurrentBib}

\bibitem [\protect \citeauthoryear {%
Zhao%
, Zhou%
, Adali%
, Zhang%
\BCBL {}\ \BBA {} Cichocki%
}{%
Zhao%
\ \protect \BOthers {.}}{%
{\protect \APACyear {2013}}%
{\protect \APACexlab {{\protect \BCnt {2}}}}}]{%
ZhaoKTD2013}
\APACinsertmetastar {%
ZhaoKTD2013}%
\begin{APACrefauthors}%
Zhao, Q.%
, Zhou, G.%
, Adali, T.%
, Zhang, L.%
\BCBL {} Cichocki, A.%
\end{APACrefauthors}%
\unskip\
\newblock
\APACrefYearMonthDay{2013{\protect \BCnt {2}}}{}{}.
\newblock
{\BBOQ}\APACrefatitle {Kernelization of Tensor-Based Models for Multiway Data
  Analysis: Processing of Multidimensional Structured Data} {Kernelization of
  tensor-based models for multiway data analysis: Processing of
  multidimensional structured data}.{\BBCQ}
\newblock
\APACjournalVolNumPages{IEEE Signal Processing Magazine}{30}{4}{137-148}.
\newblock

\newblock

\PrintBackRefs{\CurrentBib}

\bibitem [\protect \citeauthoryear {%
Zhou%
, Li%
\BCBL {}\ \BBA {} Zhu%
}{%
Zhou%
\ \protect \BOthers {.}}{%
{\protect \APACyear {2013}}%
}]{%
Zhou13}
\APACinsertmetastar {%
Zhou13}%
\begin{APACrefauthors}%
Zhou, H.%
, Li, L.%
\BCBL {} Zhu, H.%
\end{APACrefauthors}%
\unskip\
\newblock
\APACrefYearMonthDay{2013}{}{}.
\newblock
{\BBOQ}\APACrefatitle {Tensor Regression with Applications in Neuroimaging Data
  Analysis} {Tensor regression with applications in neuroimaging data
  analysis}.{\BBCQ}
\newblock
\APACjournalVolNumPages{Journal of the American Statistical
  Association}{108}{502}{540-552}.
\newblock

\newblock

\PrintBackRefs{\CurrentBib}

\end{thebibliography}
\end{document}